\documentclass[sigconf]{acmart}
\usepackage{amsmath}
\usepackage{booktabs}
\usepackage{upgreek}
\usepackage[aboveskip=0.1pt,belowskip=0.1pt]{caption}
\usepackage[labelformat=simple]{subcaption}
\usepackage{graphicx}
\usepackage{multirow}
\usepackage{enumitem}
\usepackage[ruled,vlined,linesnumbered]{algorithm2e}

\definecolor{green}{rgb}{0.0, 0.7, 0.0}
\definecolor{salmon}{rgb}{1.0,0.6,0.6}
\definecolor{yellow}{rgb}{0.88,0.74,0.13}
\definecolor{royalblue}{rgb}{0.27, 0.44, 0.78}
\definecolor{darkorange}{rgb}{0.93, 0.38, 0.17}

\let\oldnl\nl
\newcommand{\nonl}{\renewcommand{\nl}{\let\nl\oldnl}}

\AtBeginDocument{%
  \providecommand\BibTeX{{%
    \normalfont B\kern-0.5em{\scshape i\kern-0.25em b}\kern-0.8em\TeX}}}

\copyrightyear{2021}
\acmYear{2021}
\setcopyright{iw3c2w3}
\acmConference[WWW '21]{Proceedings of the Web Conference 2021}{April 19--23, 2021}{Ljubljana, Slovenia}
\acmBooktitle{Proceedings of the Web Conference 2021 (WWW '21), April 19--23, 2021 ,Ljubljana, Slovenia}
\acmPrice{}
\acmDOI{10.1145/3442381.3449974}
\acmISBN{978-1-4503-8312-7/21/04}

\setlist[enumerate]{leftmargin=1.5em}
\setlist[itemize]{leftmargin=1.5em}
\begin{document}

\title{Self-Supervised Hyperboloid Representations from Logical Queries over Knowledge Graphs}

\author{Nurendra Choudhary$^1$, Nikhil Rao$^2$, Sumeet Katariya$^2$, Karthik Subbian$^2$, Chandan K. Reddy$^{1,2}$}
\affiliation{\institution{$^1$Department of Computer Science, Virginia Tech, Arlington, VA, USA} \city{} \country{}}
\affiliation{\institution{$^2$Amazon, Palo Alto, CA, USA} \city{} \country{}}
\email{nurendra@vt.edu, {nikhilsr, katsumee, ksubbian}@amazon.com, reddy@cs.vt.edu}

\renewcommand{\shortauthors}{Choudhary et al.}
\newcommand\superequiv{\mathrel{\rlap{\raisebox{\fontdimen22\textfont2}{$=$}}\raisebox{-0.5\fontdimen22\textfont2}{$ = $}}}
\newcommand{\nuren}[1]{{\color{blue} {\sc} {#1}}}
\newcommand{\nr}[1]{{\color{red} {\sc} {#1}}}
\newcommand{\sk}[1]{{\color{green} {\sc} {#1}}}

\begin{abstract}
	Knowledge Graphs (KGs) are ubiquitous structures for information
	storage in several real-world applications such as web search, e-commerce,
	social networks, and biology. Querying KGs remains a foundational and challenging problem due to their size and complexity. Promising approaches to tackle
	this problem include embedding the KG units (e.g., entities and relations)
	in a Euclidean space such that the query embedding contains
	the information relevant to its results. These approaches, however,
	fail to capture the hierarchical nature and semantic information of the
	entities present in the graph. Additionally, most of these approaches
	only utilize multi-hop queries (that can be modeled by simple translation operations) to learn embeddings and ignore more complex operations such as intersection, and union of simpler queries. To tackle such complex operations, in this paper, we
	formulate KG representation learning as a self-supervised logical
	query reasoning problem that utilizes translation, intersection and
	union queries over KGs. We propose Hyperboloid Embeddings
	(HypE), a novel self-supervised dynamic reasoning framework, that
	utilizes positive first-order existential queries on a KG to learn representations
	of its entities and relations as hyperboloids in a Poincaré
	ball. HypE models the positive first-order queries as geometrical
	translation, intersection, and union. For the problem of KG reasoning in real-world datasets, the proposed HypE model significantly
	outperforms the state-of-the art results. We also apply HypE to an
	anomaly detection task on a popular e-commerce website product taxonomy as well as hierarchically organized web articles and demonstrate significant performance improvements compared to existing baseline methods. Finally, we also visualize the learned HypE embeddings in a Poincaré ball to clearly interpret and comprehend
	the representation space.



\end{abstract}

\ccsdesc[500]{Computing methodologies~Reasoning about belief and knowledge}
\ccsdesc[300]{Computing methodologies~Knowledge representation and reasoning}
\ccsdesc[100]{Computing methodologies~Representation of mathematical objects}

\keywords{Representation learning, knowledge graphs, hyperbolic space, reasoning queries}

%

\maketitle

\section{Introduction}
\begin{figure}[htbp]
	\vspace{-1.5em}
	\centering
	\includegraphics[width=\linewidth]{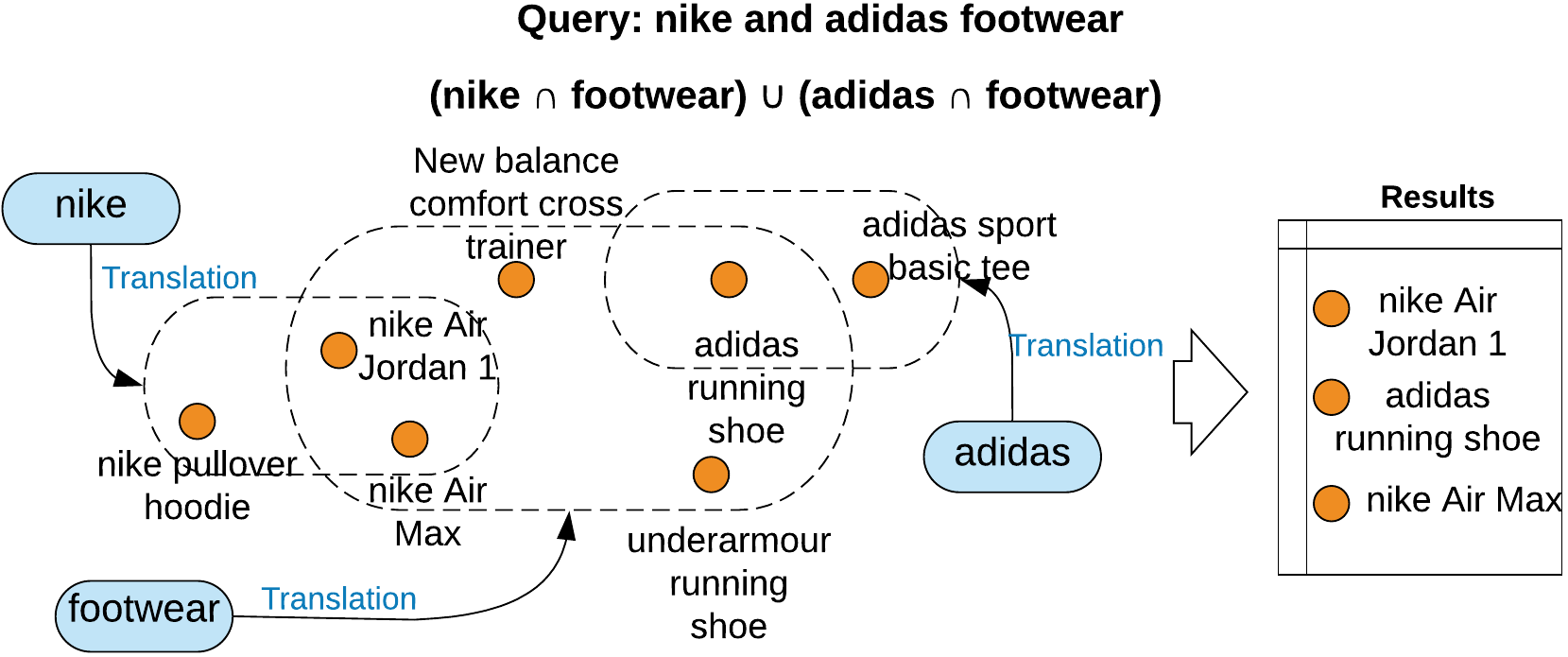}
	\caption{An example of PFOE querying in the E-commerce product network. The product space of \texttt{Adidas} and \texttt{Nike} intersects with \texttt{Footwear} to narrow the search space. A union over these spaces yields our final set of entity results.}
	\label{fig:pfoe}
	\Description[PFOE Example]{An example that shows how a first-order query can provide its results from a Knowledge Graph.}
	\vspace{-1.5em}
\end{figure}

Knowledge Graphs (KGs) organize information as a set of entities connected by relations. Positive first-order existential (PFOE) queries such as translation, intersection, and union over these entities aid in effective information extraction from massive data (see Figure~\ref{fig:pfoe} for an example PFOE query). Efficient handling of such queries on KGs is of vital importance in a range of real-world application domains including search engines, dialogue systems, and recommendation models. However, the large size of KGs and high degrees of the nodes therein makes traversal for querying a computationally challenging or, in some cases, even an impossible task \cite{wang2017knowledge}. 
One way to resolve this issue is to learn representations for the KG units (entities and relations) in a latent (generally Euclidean) space such that algebraic or logical operations can be applied to extract relevant entities. Robust representation learning of KG units has several real-world applications including KG information extraction \cite{hamilton2018embedding}, entity classification \cite{wilcke2020endtoend}, and anomaly detection \cite{8455737}.

Earlier approaches to representation learning in KGs model entities and relations as vectors in the Euclidean space \cite{NIPS2013_5071,nickel2015holographic,yang2014embedding}. This is suboptimal due to the constant size of a point's answer space which does not capture the variations induced by different queries. Specifically, broad queries (\texttt{Nike}) should intuitively cover a larger region of the answer space compared to specific queries (\texttt{Nike running shoes for men}). In the recently proposed Query2Box model \cite{ren2020query2box}, the authors demonstrated the  effectiveness of complex geometries (such as hyper-rectangles) with varying offsets that control the size of an answer space according to a query's complexity. However, such architectures lack the ability to capture hierarchical information that is prevalent in many KGs. Furthermore, previous representation learning methods in heterogeneous graphs (or KGs) \cite{lin2018multi,zhang2020graph,liu2020k,fu2020magnn} solely focus on one-hop or multi-hop reasoning over relations. Such frameworks enable static and optimized computational graphs, but lead to poor retrieval from complex intersection and union queries. Dynamic computational graphs, which are able to modify their network architecture with a switch mechanism (discussed in Section \ref{sec:implementation}) can significantly alleviate this problem.

Although Euclidean spaces have proven to be effective for {representation learning in various domains \cite{bengio2013representation}}, several hierarchical datasets (including graph data) in the fields of network sciences and E-commerce taxonomies demonstrate a latent {non-Euclidean} anatomy \cite{bronstein2017geometric}. 
The introduction of hyperbolic algebraic operations \cite{ganea2018hyperbolic} 
have led to the proliferation of hyperbolic neural networks such as Hyperbolic-GCN (H-GCN) \cite{chami2019hyperbolic} and Hyperbolic Attention (HAT) networks \cite{gulcehre2018hyperbolic}. These frameworks leverage the hyperbolic anatomy of hierarchical datasets and show a significant performance boost compared to their Euclidean counterparts. To the best of our knowledge, there is no existing work that (i) utilizes dynamic computational graphs on the hyperbolic space, (ii) applies complex hyperbolic geometries such as hyperboloids for representation learning. Additionally, the static computational graphs of H-GCN and HAT limit their learning capability to a single problem, generally, multi-hop (translation) reasoning. This severely limits their applicability to representation learning on KGs since translations can only utilize single entities. \textbf{More complex intersections and unions not only use more entities, but are also more representative of real-world KG queries.} While solving union and intersection queries is more challenging, they enable better representation learning \cite{DBLP:journals/corr/abs-1206-5538}. Traversing over the entities in KGs facilitates an intuitive way of constructing a query-reasoning proxy task (refer Section \ref{sec:query}) that enables representation learning of entities and relations. These representations, in a self-supervised framework, can further provide enriched features in downstream tasks with smaller annotated datasets (such as anomaly detection), thus alleviating the issue of data scarcity.

\begin{figure}[htbp]
	\centering
	\begin{subfigure}[b]{.49\linewidth}
		\centering
		\includegraphics[width=\linewidth]{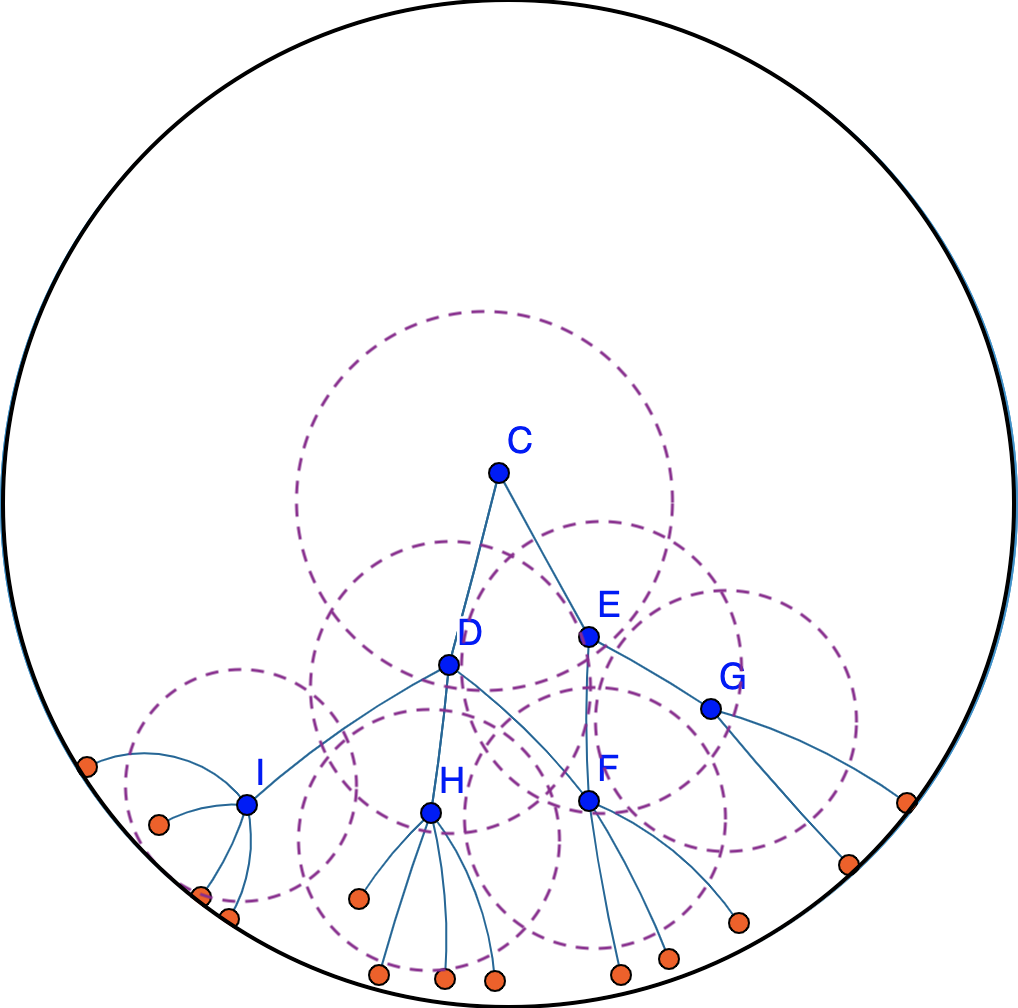}
		\caption{Hyperbolic vectors}
		\label{fig:hyperbolic_vectors}
	\end{subfigure}
	\begin{subfigure}[b]{.49\linewidth}
		\centering
		\includegraphics[width=\linewidth]{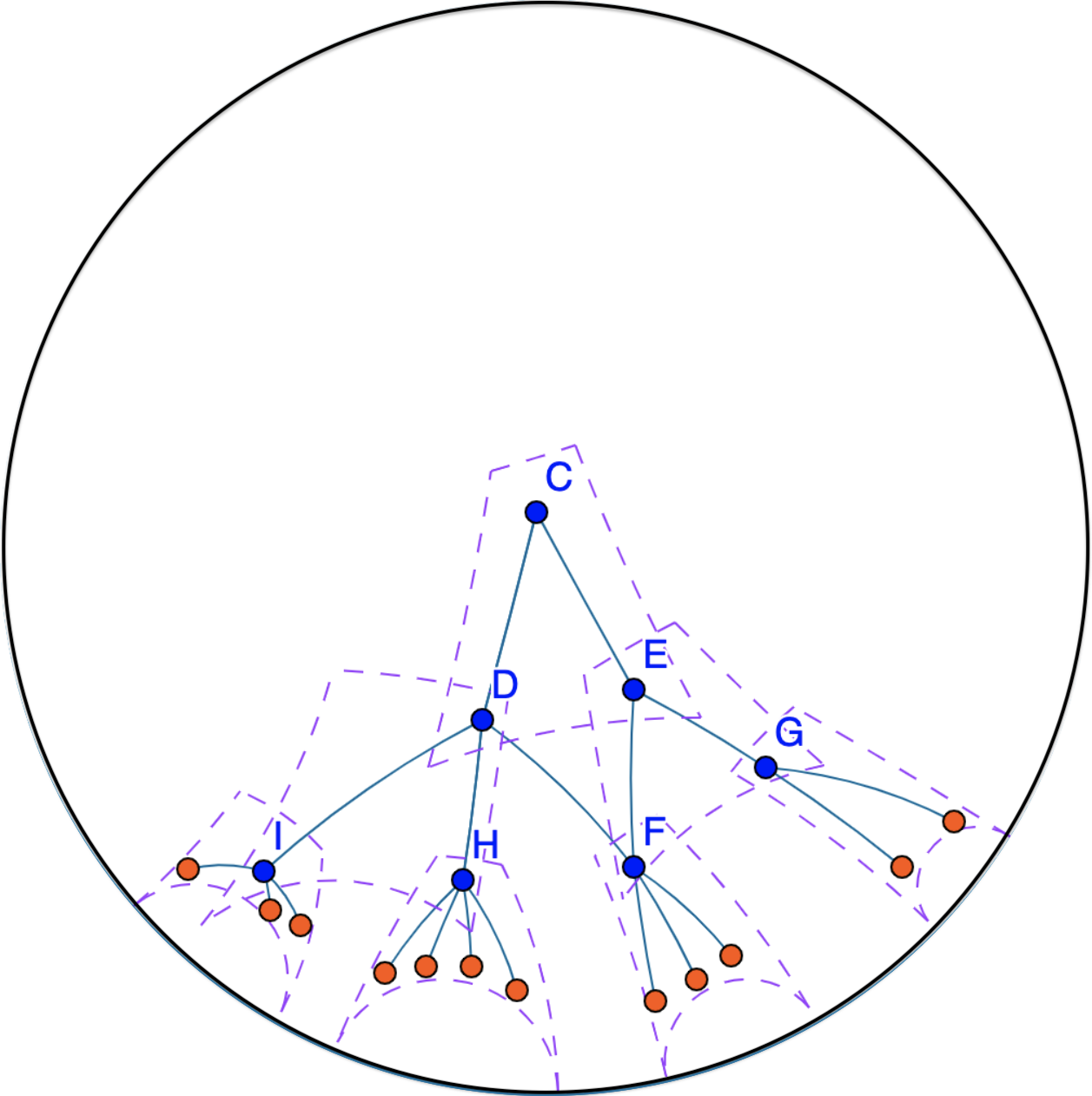}
		\caption{Hyperboloids}
		\label{fig:hyperboloids_dynamic}
	\end{subfigure}
	\caption{Static vs Dynamic representation. (a) hyperbolic vectors have lower precision due to static thresholds over a center, depicted by the dotted circle. (b) Dotted hyperboloids encapsulate all its child entities because of dynamic sizes. The {\color{blue}{blue}} and {\color{orange}{orange}} circles are intermediate and leaf nodes.}
	\Description[Hyperbolic vectors and Hyperboloids]{Visualization of hyperbolic vectors and hyperboloids in Poincaré ball space.}
	\label{fig:dynamic_limit}
	\vspace{-1.5em}
\end{figure}
Motivated by the effectiveness of self-supervised learning and the need for non-Euclidean geometries in KGs, we formulate KG representation learning as a self-supervised query reasoning problem. \textit{We introduce Hyperboloid Embeddings (HypE), a self-supervised dynamic representation learning framework that utilizes PFOE queries to learn hyperboloid representations of KG units in a (non-Euclidean) Poincaré hyperball.} Hyperboloids, unlike vectors in hyperbolic spaces, allow us to use dynamic sizes for KG representations. 
For e.g., in Figure \ref{fig:dynamic_limit}, we can notice that different entities contain different number of children, and, thus learning a static vector representation is suboptimal. Hyperboloids learn an additional spatial parameter, \textit{limit} (described in Section \ref{sec:reasoning}), that can model the varying entity sizes. Moreover, the dynamic nature of its computational graphs allows HypE to utilize different network layers to learn different types of operations, namely, translation, intersection, and union; and process varying number of input units depending on the learning operation. Our empirical studies include learning representations from large-scale KGs in e-commerce, web pages (DBPedia), and other widely used Knowledge bases (such as Freebase and NELL995); and evaluating the representations on the downstream task of 
anomaly detection. The major contributions of this paper are:
\begin{enumerate}
	\item {Formulate} the KG representation learning problem as a self-supervised query reasoning problem to leverage PFOE queries.
	\item {Introduce} Hyperboloid Embeddings (HypE), a self-supervised dynamic representation learning framework that learns hyperboloid representations of KG units in a Poincaré hyperball. This is motivated by the need for non-Euclidean geometries.
	\item {Perform} an extensive set of empirical studies across diverse set of real-world datasets to evaluate the performance of HypE against several state-of-the-art baseline methods on the  downstream task of {Anomaly Detection.}
	\item {Visualize} the HypE embeddings to clearly interpret and comprehend the representation space. 
\end{enumerate}
The rest of the paper is organized as follows: Section \ref{sec:related} describes the related background. Section \ref{sec:hyperboloid} formulates the representation learning problem, explains the non-Euclidean algebraic operations, and the proposed HypE model. In Section \ref{sec:experimental}, we describe the real-world datasets, state-of-the-art baselines and performance metrics used to evaluate the HypE model. We demonstrate the performance results along with the visualization of HypE's representations. Finally, Section \ref{sec:conclusion} concludes the paper. 

\section{Related Work}
\label{sec:related}
In this section, we review different geometries utilized for learning representations and earlier works that are adopted for reasoning over Knowledge Graphs.
\paragraph{Representation Geometries :}
Previous approaches to representation learning, in the context of KG, aim to learn latent representations for entities and relations. 
Translational frameworks \cite{NIPS2013_5071,nickel2011three,yang2014embedding} model relations using translation between entity pairs. This limits the models to only {handle translation-based} queries. Graph Query Embedding (GQE) \cite{hamilton2018embedding} overcame this limitation and provided a technique for leveraging intersection queries as deep sets \cite{zaheer2017deep} over different queries. Furthermore, Box Lattices \cite{vilnis2018probabilistic}, EMQL \cite{sun2020faithful} and Query2Box \cite{ren2020query2box} proved the effectiveness of more complex geometries (hyper-rectangles) for queries. Word2Gauss \cite{vilnis2014word} is a popular NLP technique that learns Gaussian embeddings for words. DNGE \cite{pei2019dynamic} utilizes a dynamic network to learn Gaussian embeddings for entities in a graph. 
 These Gaussian representations cannot be intuitively extended to KGs because they are not closed under more complex PFOE queries (intersection or union of Gaussians does not yield a Gaussian). Furthermore, they rely on properties of the Euclidean space to learn representations, which are proven ineffective at capturing the prevalent hierarchical features of a KG \cite{ganea2018hyperbolic}.

\paragraph{{Representation Learning on Graphs :}}
One of the fundamental problems in KG is to aggregate the neighbor information of nodes while learning representations.  Node embedding techniques such as Node2Vec \cite{grover2016node2vec} and DeepWalk \cite{perozzi2014deepwalk} aggregate the neighbors' features by modeling the node's dependence on its neighbors. ChebNet \cite{defferrard2016convolutional} uses Chebyshev ploynomials and filters node features in the graph Fourier domain. GCN \cite{kipf2016semi} constrains the parameters of ChebNet to alleviate overfitting and shows improved performance. Graph-BERT \cite{zhang2020graph} and MAGNN \cite{fu2020magnn} provide a self-supervised learning model utilizing the tasks of masking and metapath aggregation, respectively. 
In another line of research, \citeauthor{miller2009nonparametric} \cite{miller2009nonparametric} utilizes non-parametric Bayesian frameworks for link prediction on social networks. \citeauthor{zhu2016max} \cite{zhu2016max} further improved {the approach with a max-margin framework.} KGAT \cite{wang2019kgat} is another popular approach that utilizes attention networks over entities and relations with a TransR \cite{lin2017learning} loss function to learn representations for user recommendation. These methods rely on relational properties and thus are effective in handling translational problems such as multi-hop reasoning. However, they are ineffective at handling more complex PFOE queries such as intersection and union.

Other popular multi-hop graph networks such as Graph Attention Network (GAT) \cite{velivckovic2017graph}  and Graph Recurrent Network (GRN) \cite{song2018graph} have previously shown impressive results in reasoning-based QA tasks. 
However, hyperbolic flavors of these networks, H-GNN \cite{ganea2018hyperbolic}, H-GCN \cite{chami2019hyperbolic,chami-etal-2020-low} and H-GAT \cite{gulcehre2018hyperbolic} argue that hierarchical datasets follow the anatomy of hyperbolic space and show improved performance over their Euclidean counterparts. {Nonetheless, these approaches are still limited by the constant answer space that does not consider the varying fluctuations of complex queries.} 

Self-supervised learning \cite{doersch2017multi, jamaludin2017self, xu2019ternary, nagrani2020disentangled} 
utilizes large unannotated datasets to learn representations that can be fine-tuned to other tasks that have relatively smaller amount of annotated data. 
{Traversing over the entities in KGs facilitates an intuitive way of constructing a query-reasoning proxy task (refer Section \ref{sec:query}) that enables representation learning of entities and relations. These representations, in turn, are employed in downstream tasks with scarce datasets such as anomaly detection.}

{The proposed HypE model utilizes a self-supervised learning framework that leverages both simple and complex PFOE queries to learn hyperboloid {(with varying limits)} representations of KG units in a Poincaré ball to efficiently capture hierarchical information.}

\section{Proposed Framework}
\label{sec:hyperboloid}
{In this section, we first provide the standard method of querying knowledge graphs. Then, we set up the problem and describe the details of our model that learns representations of entities and relations from reasoning queries over Knowledge Graphs (KG).

\subsection{Querying Knowledge Graphs}
KGs contain two primary units, namely, entities and relations. Entities are the basic information units that connect to each other by relation units. Heterogeneous graphs \cite{wang2019heterogeneous,10.1145/3292500.3330961} can be considered as a special case of KGs 
 where the relations serve as hierarchical connections with no inherent information. PFOE queries of translation (t), intersection ($\cap$) and union  ($\cup$) serve as the primary means of querying these KGs. Translation queries utilize an entity $e$ and a relation $r$ to retrieve all entities that are connected to $e$ through $r$. An equivalent example of translation query for heterogeneous graphs is to retrieve all children of a node $e$ connected by a certain edge type $r$. Intersection and Union operate over multiple entities $E = \{e_1, .., e_n\}$ and correspondingly retrieve the set of all entities that are connected to all $e_i \in E$ and any $e_i \in E$. For heterogeneous graphs, the equivalent is to retrieve nodes connected to all nodes $e_i \in E$ and any $e_i \in E$. An example of PFOE querying is given in Figure \ref{fig:query}. The widely studied problem of multi-hop traversal \cite{fu2020magnn} is a more specific case of translation queries, where multiple queries are chained in a series. 
\begin{figure}[htbp]
		\vspace{-1.4em}
	\includegraphics[width=\linewidth]{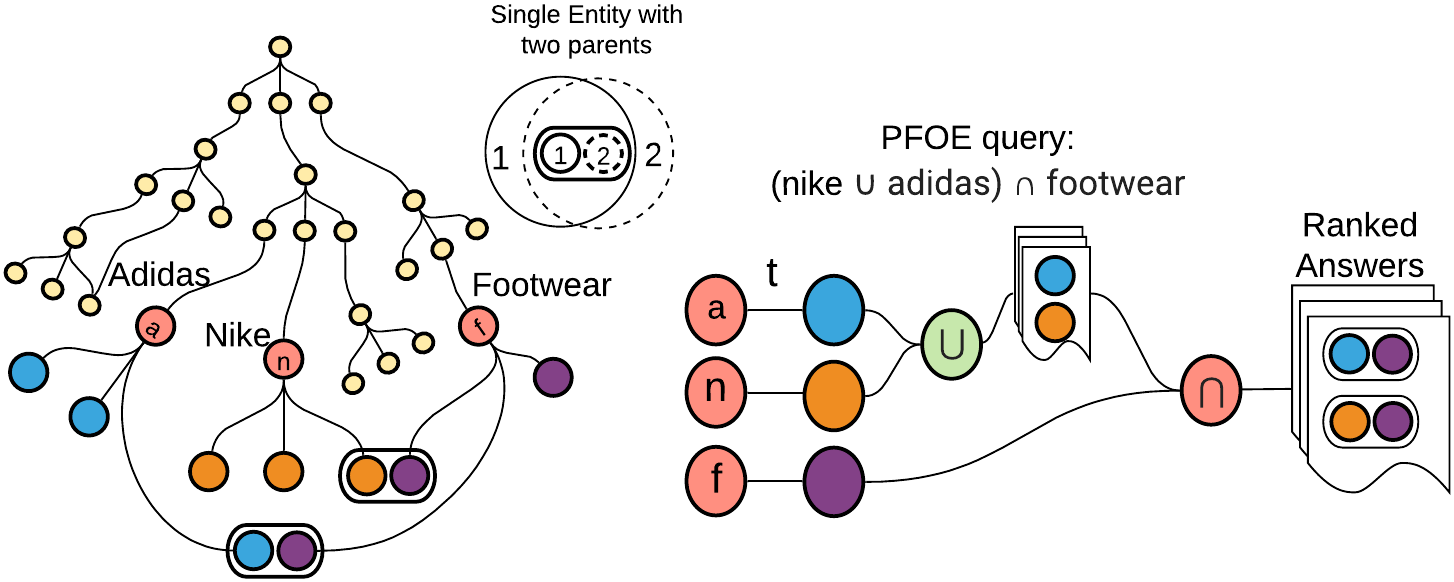}
	\caption{A simple first-order query over a Product Graph. The query \texttt{Nike and Adidas footwear} can be expressed as the union of the Nike and Adidas nodes intersected with the node corresponding to footwear in the product catalog.}
	\Description[First-order query]{Figure shows the product graph and visualizes each node's connect to an actual query processing on the knowledge graph.}
	\label{fig:query}
		\vspace{-1.8em}
\end{figure}}

\subsection{Problem Setup}
\label{sec:problem}
We denote $KG = (E,R)$ as a set of entities $e_i \in E$  and relations $r_{ij} \in R ~\ : ~\ e_i \rightarrow e_j$ as Boolean functions that indicate whether a directed relation $r_{ij}$ holds between $e_i$ and $e_j$. {Intersection ($\cap$) and Union ($\cup$)} are {positive first-order existential (PFOE)} operations defined on a set of queries $q_1, q_2, ...., q_n \in Q$:  
\begin{align*}
	q_{\cap}[Q] &= V_{\cap}  \subseteq E \mkern9mu\exists \mkern9mu e_1,e_2, ..., e_k:q_1 \cap q_2 \cap q_3 ... \cap q_n \\
	q_{\cup}[Q] &= V_{\cup}  \subseteq E \mkern9mu\exists\mkern9mu e_1,e_2, ..., e_k:q_1 \cup q_2 \cup q_3 ... \cup q_n 
\end{align*}
where $(q_{\cap},V_{\cap})$ and $(q_{\cup},V_{\cup})$ is the query space and resultant entity set after the {intersection} and {union} operations over query set $Q$, respectively\footnote{{e.g., if $(q_i,\{e_a,e_b\})$ and $(q_j,\{e_b,e_c\})$ are the corresponding query space and resultant entity sets, then $q_{\cap}[\{q_i,q_j\}] = \{e_b\}$ and $q_{\cup}[\{q_i,q_j\}]= \{e_a,e_b,e_c\}$.}}. 
Given a dataset of logical queries over a $KG$, the goal, here, is to learn hyperboloid (in a Poincaré ball) representations for its entities $e_i \in \mathbb{R}^{2d}$ and relations $r_{ij} \in \mathbb{R}^{2d}$, where $d$ is a hyper-parameter that defines the dimension of the Poincaré ball. 

\subsection{Manifold Transformation Layer}
\label{sec:poincare}
{Hierarchical structures intuitively demonstrate the latent characteristics of a hyperbolic space \cite{ganea2018hyperbolic}. Thus, we utilize the Poincaré ball \cite{cannon1997hyperbolic} to model our representations.} 

\subsubsection{Transformation from Euclidean Space} 
The transformation from Euclidean to hyperbolic space $(\mathbb{H}^n,g^\mathbb{H})$, given in \cite{ganea2018hyperbolic},  
is defined by the manifold $\mathbb{H}^n=\{x\in\mathbb{R}^n:\| x \| <1\}$ {with} the Reimannian metric , $g^\mathbb{H}$, where:

\begin{equation}
\label{eq:transformation}
	g_x^\mathbb{H}=\lambda_x^2 ~\ g^{\mathbb{E}} \quad \text{where }\lambda_x\coloneqq\frac{2}{1-\| x \|^2} 
\end{equation}
$g^\mathbb{E}=\textbf{I}_n$ being the Euclidean identity metric tensor, and $\| x \|$ is the Euclidean norm of $x$. $\lambda_x$ is the conformal factor between the Euclidean and hyperbolic metric tensor set to a conventional curvature of -1. Eq. \eqref{eq:transformation} allows us to convert a Euclidean 
metric to  hyperbolic. Thus, the distance between points $x,y \in \mathbb{H}^n$ is derived as:  
\begin{equation}
	d_\mathbb{H}(x,y) = \cosh^{-1}\left(1+2\frac{\| x-y\| ^2}{\left(1-\| x\| ^2\right)\left(1-\| y\| ^2\right)}\right)
\end{equation}

\subsubsection{Gyrovector Spaces}
Algebraic operations such as addition and scalar product which are straightforward in the Euclidean space cannot be directly applied in hyperbolic space. Gyrovector spaces allow for the formalization of these operations in hyperbolic space. 

\begin{figure}
	\includegraphics[width=.95\linewidth]{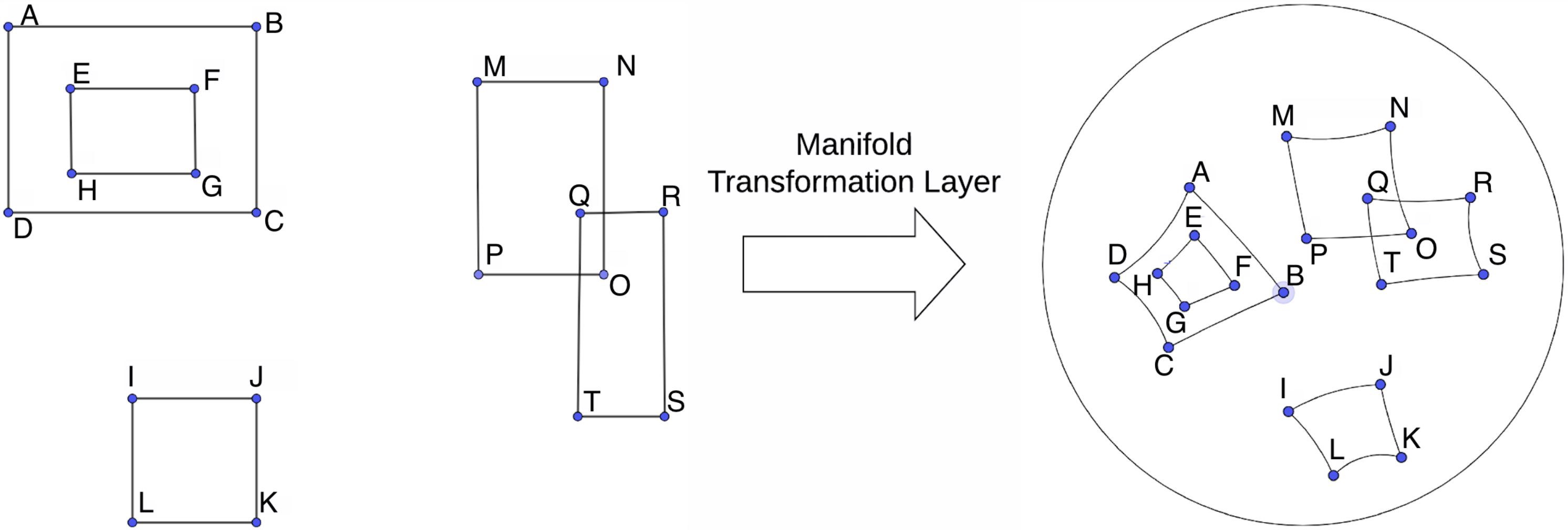}
	\caption{Manifold Transformation of Euclidean geometries (rectangles) to a Poincaré ball (horocycle enclosures).} 
	\label{fig:manifold}
	\Description[Manifolds]{The figure shows Euclidean boxes being converted to their hyperbolic counterparts or hyperboloids.}
			\vspace{-1.5em}
 \end{figure}

\citeauthor{ganea2018hyperbolic} \cite{ganea2018hyperbolic} provide the gyrovector operations relevant to training neural networks. The gyrovector operations for Poincaré ball of radius $c$ are Möbius addition $(\oplus_c)$, Möbius subtraction $(\ominus_c)$, exponential map $(\exp_x^c)$, logarithmic map $(\log_x^c)$ and Möbius scalar product $(\odot_c)$.

\begin{align*}
x \oplus_c y &\coloneqq \frac{\left(1+2c\langle x,y\rangle +c\| y\| ^2\right)x+\left(1-c\| x\| ^2\right)y}{1+2c\langle x,y\rangle +c^2\| x\| ^2\| y\| ^2}\\
x \ominus_c y &\coloneqq x \oplus_c -y	\\
\exp_x^c(v) &\coloneqq x \oplus_c \left(\tanh\left(\sqrt{c}\frac{\lambda_x^c\| v\| }{2}\right)\frac{v}{\sqrt{c}\| v\| }\right)\\
\log_x^c(y) &\coloneqq \frac{2}{\sqrt{c}\lambda_x^c}\tanh^{-1}\left(\sqrt{c}\| -x\oplus_cy\| \right)\frac{-x\oplus_cy}{\| -x\oplus_cy\| }\\
r \odot_c x &\coloneqq \exp_0^c(rlog_0^c(x)), ~\forall r \in \mathbb{R}, x \in \mathbb{H}^n_c
\end{align*}
Here, $\coloneqq$ denotes assignment operation for Möbius operations.
Also, the norm of $x, y$ can subsume the scaling factor $c$.
Hence, in HypE, training can be done with a constant $c$ or trainable $c$. We empirically validate this assumption in our experiments {(Section \ref{sec:ablation})}. 
Figure \ref{fig:manifold} shows an example of the manifold transformation from Euclidean space to a Poincaré ball of unit radius. HypE extends the operations to handle complex geometries, explained in Section \ref{sec:reasoning}. 
 
\subsection{Dynamic Reasoning  Framework : HypE}
\label{sec:reasoning}
\begin{figure}[htbp]
	\centering
	\begin{subfigure}[b]{0.49\linewidth}
		\centering
		\includegraphics[width=.8\linewidth]{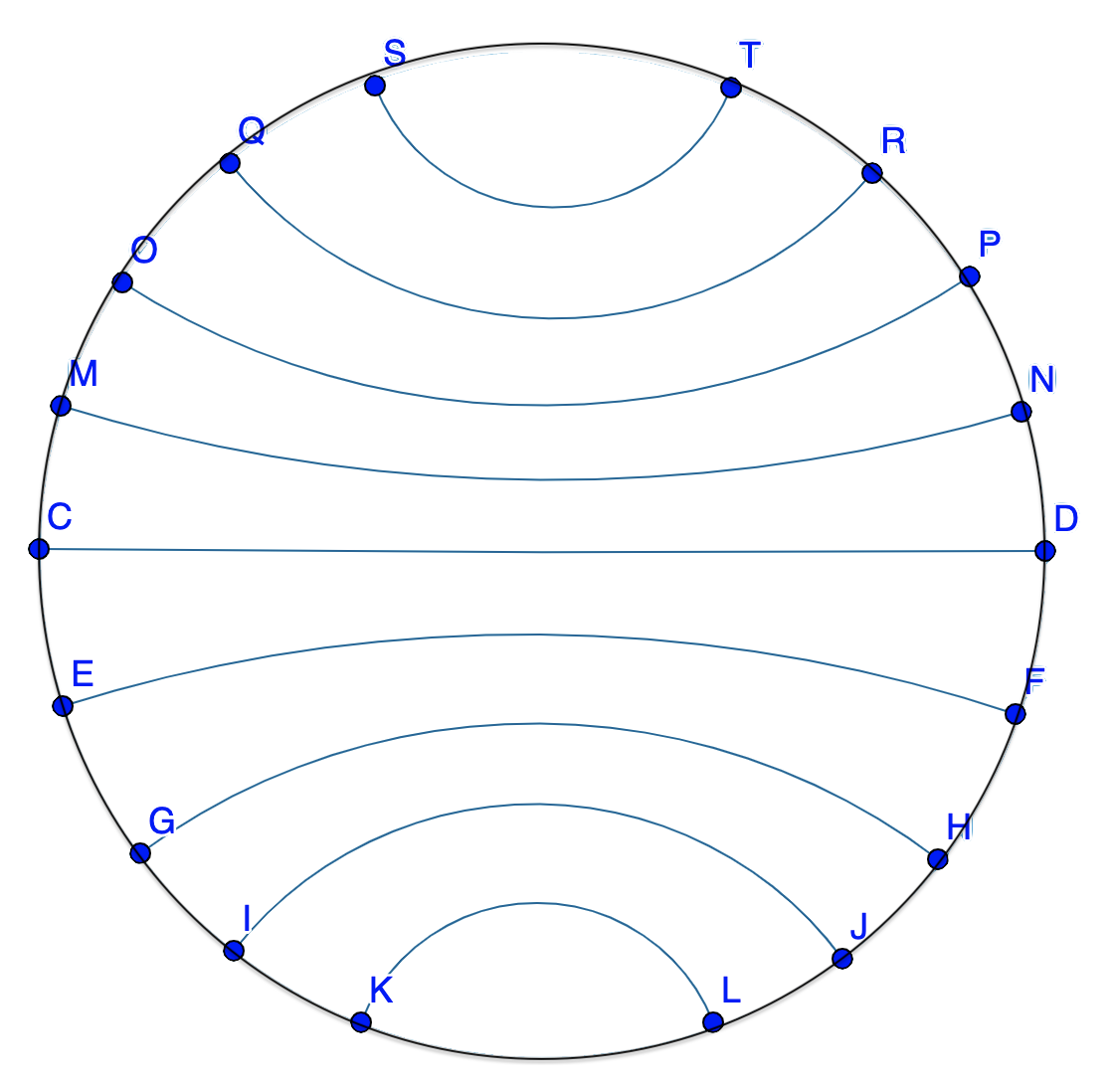}
		\caption{Horocycles in a Poincaré ball.}
		\label{fig:horocycle}
	\end{subfigure}
	\begin{subfigure}[b]{0.49\linewidth}
		\centering
		\includegraphics[width=.8\linewidth]{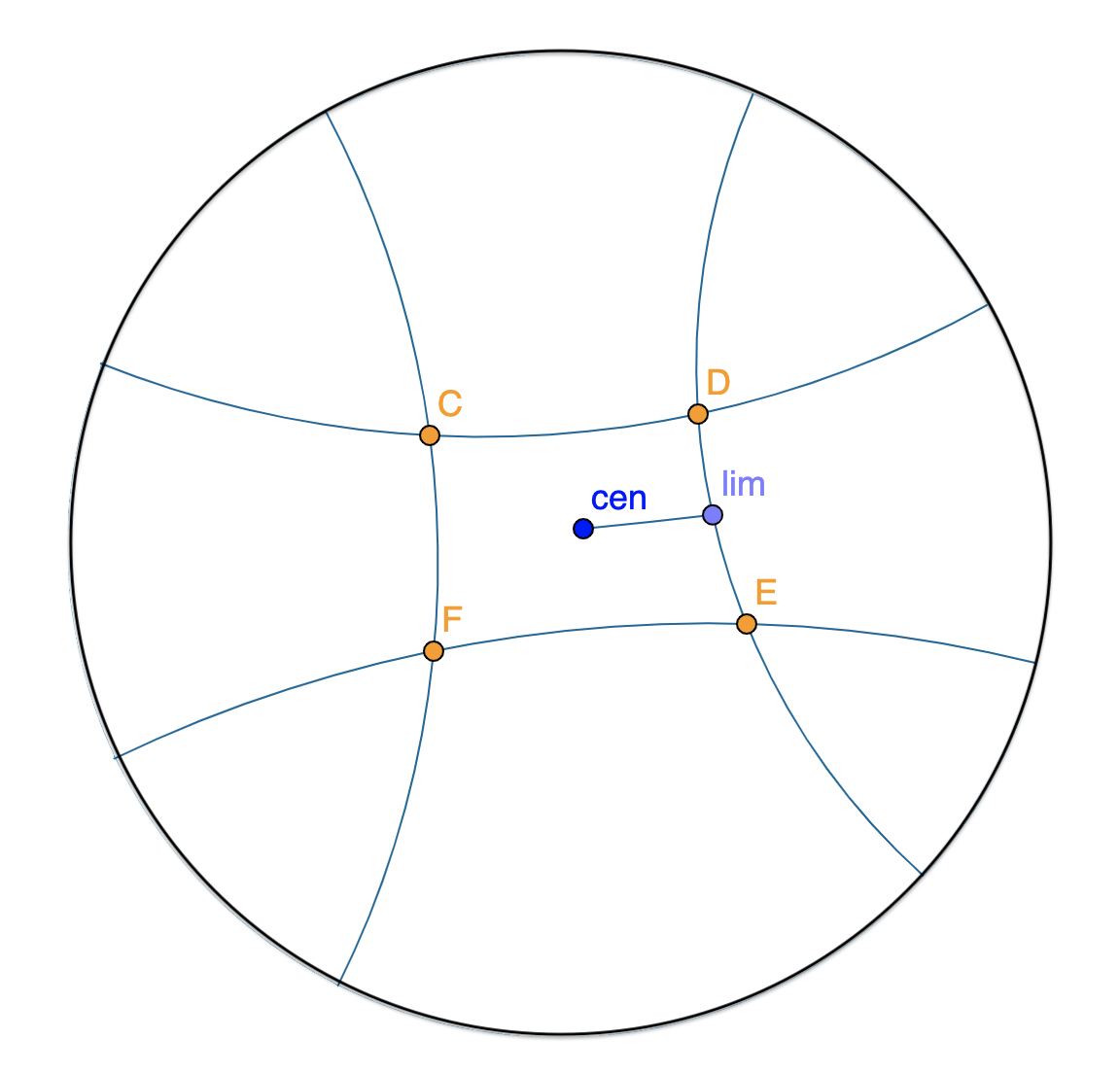}
		\caption{CDFE is the hyperboloid.}
		\label{fig:hyperboloid}
	\end{subfigure}
	\caption{Horocycles and hyperboloids in a Poincaré ball. The hyperboloid is composed of two parallel pairs of arc-aligned horocycles.}
	\Description[Horocycles and Hyperboloid]{The first figure shows a horocycle which are parallel lines in hyperbolic space and the second figure shows that their intersection leads to a hyperboloid.}
	\label{fig:representation}
			\vspace{-1.5em}
\end{figure}

\begin{figure}
	\includegraphics[width=\linewidth]{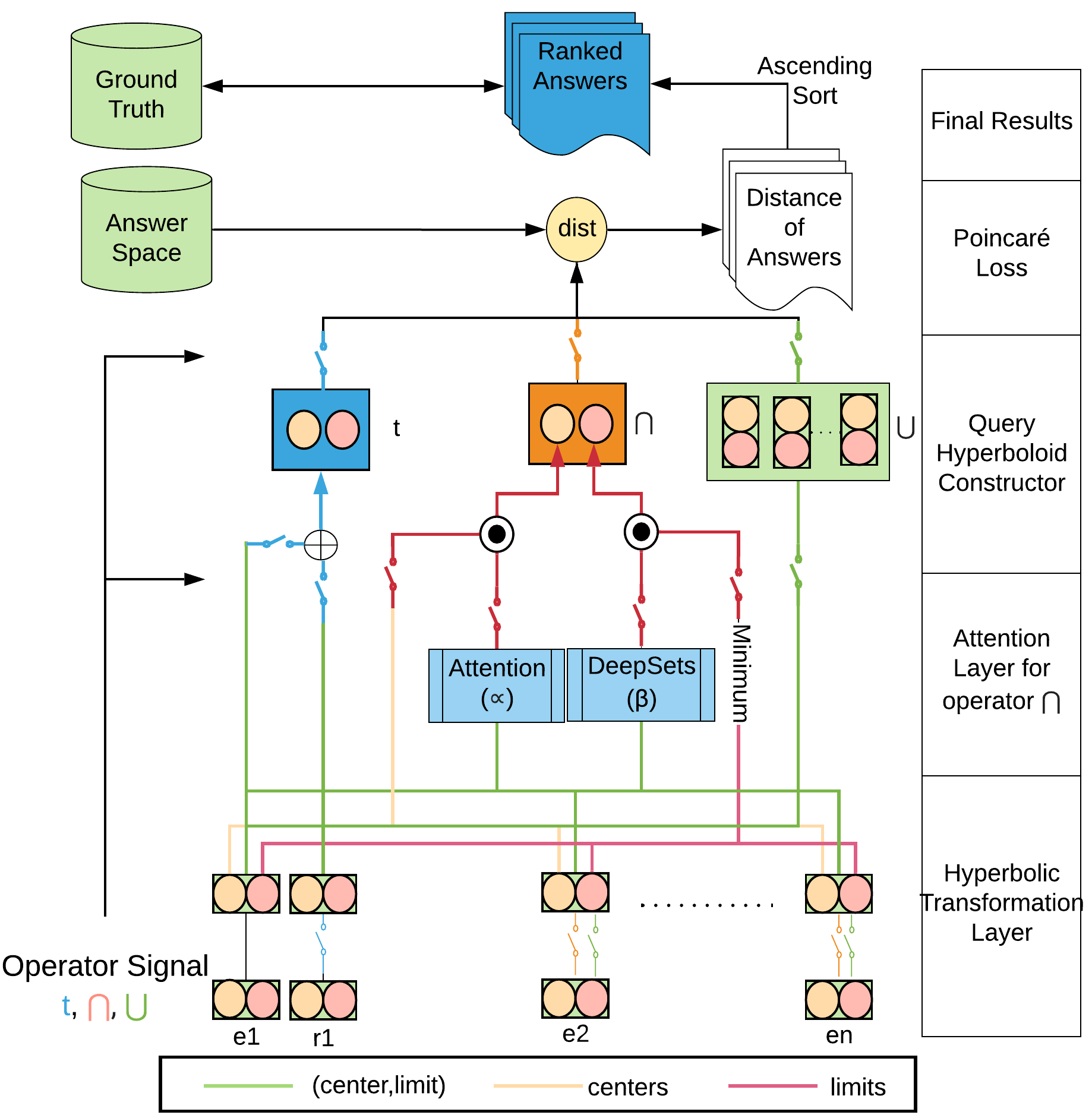}
	\caption{{An overview of the proposed HypE architecture. The architecture utilizes a switch mechanism to connect/disconnect different layers according to the query operator signal (t,$\cap$,$\cup$). The {\color{cyan}{blue}}, {\color{red}{red}} and {\color{green}{green}} switches are connected for translation, intersection and union operations, respectively (and disconnected otherwise). The {\color{yellow}yellow} and {\color{salmon}pink} circles depict the center and limit of KG units, respectively. This figure is best viewed in color. }}
	\Description[HypE]{The figure shows the connections between different units. The input is raw embeddings connected through various layers with a switch mechanism and finally it outputs the ranked results.}
	\label{fig:hype}
	\vspace{-1.5em}
\end{figure}

{We aim to learn hyperboloid (made of two parallel pairs of arc-aligned horocycles) embeddings for all the entities and relations in the KG. An arc-aligned horocyle (Figure \ref{fig:horocycle}) is a partial circle that is parallel to the diameter of a Poincaré ball and orthogonally intersects its boundaries at two points}. A hyperboloid embedding (see Figure \ref{fig:hyperboloid}) $e =\left(\text{cen}(e), \text{lim}(e)\right) \in \mathbb{R}^{2d}$ is characterized by:

\begin{equation*}
	H_e \superequiv \{v\in\mathbb{R}^d:\text{cen}(e)\ominus_c\text{lim}(e)\leq v \leq \text{cen}(e)\oplus_c\text{lim}(e)\}
\end{equation*}
where $\superequiv$ denotes strict equivalence and $\leq$ is element-wise inequality and $\text{cen}(t), \text{lim}(t) \in \mathbb{R}^d$ are center of the hyperboloid and positive border limit $(\text{lim}(t) \geq 0)$ of the enclosing horocycle from the center, respectively.  The overview of the architecture is given in Figure \ref{fig:hype}.
From the KG, we derive the following types of directed edge relations to build our dynamic computational graph for learning embeddings. 

\noindent	\textbf{Distance between hyperboloid and entity point (d):} Given a query hyperboloid $q \in \mathbb{R}^{2d}$ and entity center $v \in \mathbb{R}^d$, 
	the distance between them is defined as:
	\begin{align}
		d_{hyp}(v,q)&= d_{out}(v,q) \oplus_c \gamma d_{in}(v,q)\label{eq:hyp_dist}\\
		d_{out}(v,q)&= \| Max(d_{\mathbb{H}}(v, q_{max}), 0) + Max(d_{\mathbb{H}}(q_{min},v), 0)\| _1 \nonumber\\
		d_{in}(v,q)&= \| cen(q) \ominus_c Min(q_{max}, Max(q_{min}, v))\| _1 \nonumber\\
		q_{min}&= cen(q) \ominus_c lim(q), \quad q_{max} = cen(q) \oplus_c lim(q) \nonumber
	\end{align}
	where $d_{out}$ represents the distance of the entity to limits of the hyperboloid and $d_{in}$ is the distance of the entity from {the hyperboloid's border to its center.} $\gamma$ is a scalar weight (set to 0.5 in our experiments) and $\| x\| _1$ is the $L1$-norm of x.  
	
\begin{figure}
		\centering
		\begin{subfigure}[b]{0.49\linewidth}
			\centering
			\includegraphics[width=.84\linewidth]{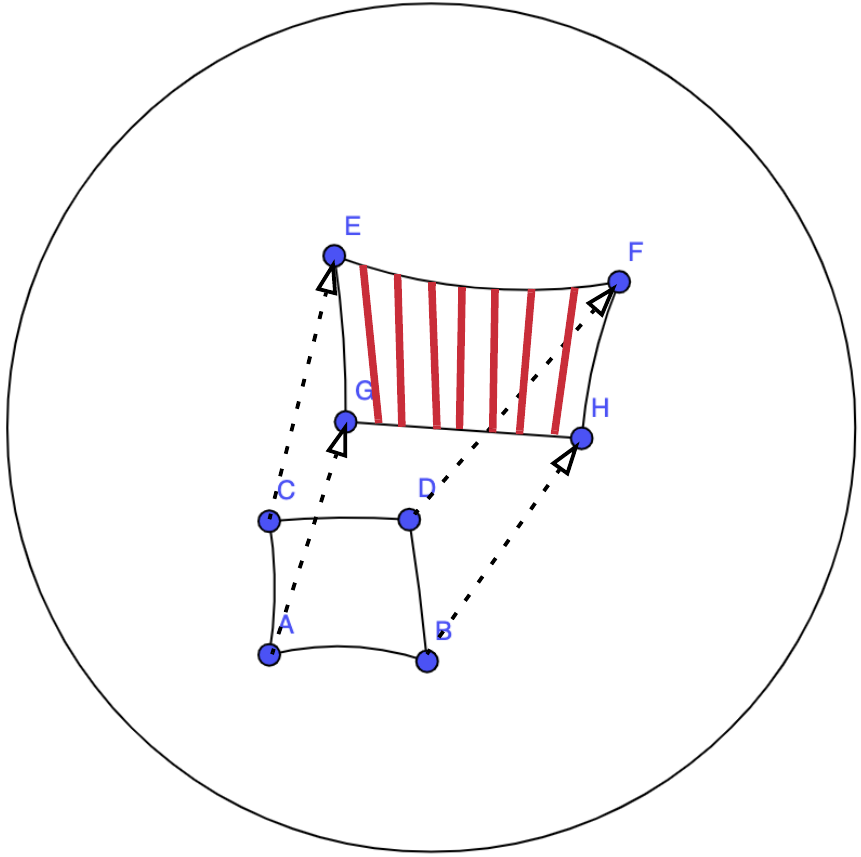}
			\caption{Translation}
			\label{fig:translation}
		\end{subfigure}
		\begin{subfigure}[b]{0.49\linewidth}
			\centering
			\includegraphics[width=.84\linewidth]{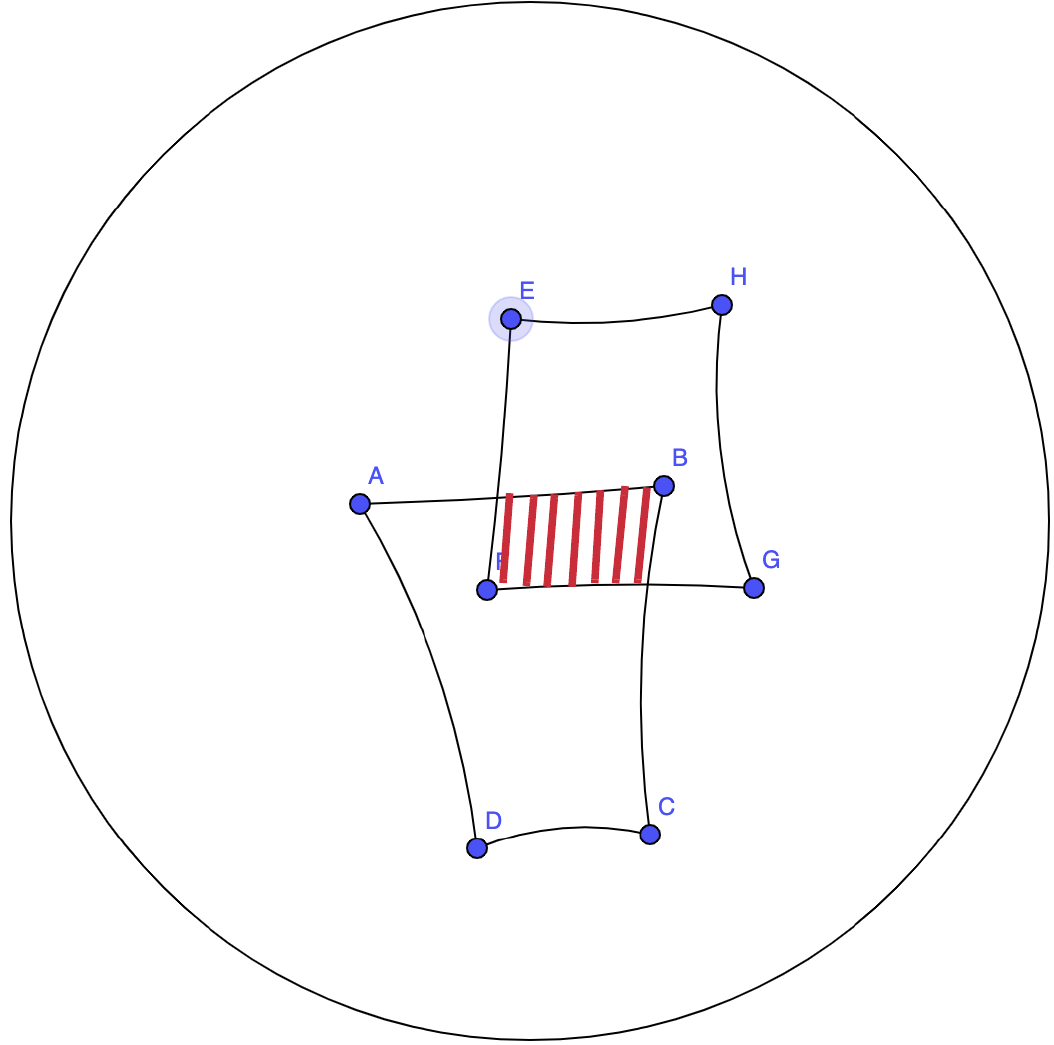}
			\caption{Intersection}
			\label{fig:intersection}
		\end{subfigure}
			\begin{subfigure}[b]{0.49\linewidth}
		\centering
		\includegraphics[width=.84\linewidth]{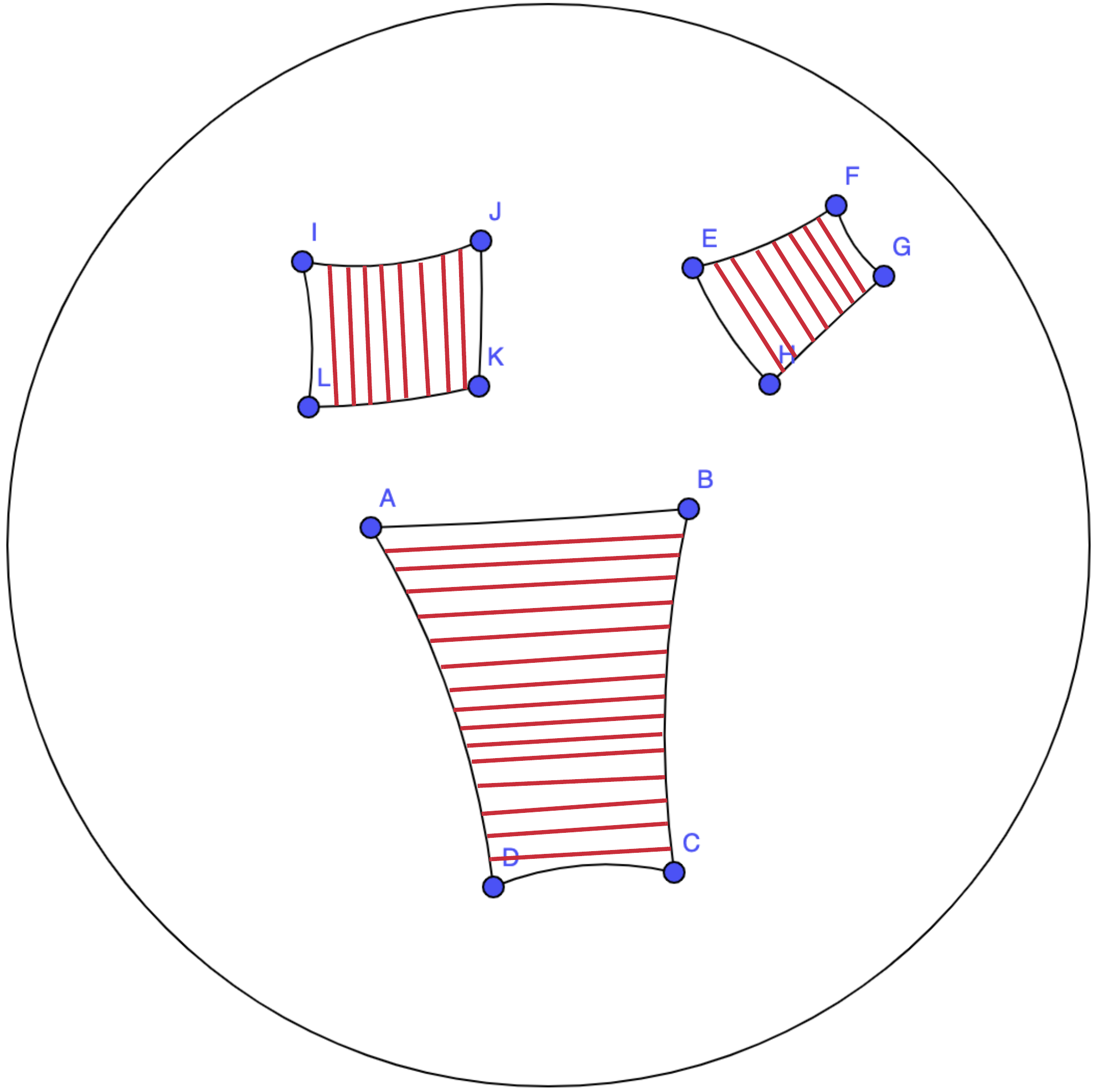}
		\caption{Union}
		\label{fig:union}
	\end{subfigure}
	\begin{subfigure}[b]{0.49\linewidth}
		\centering
		\includegraphics[width=.84\linewidth]{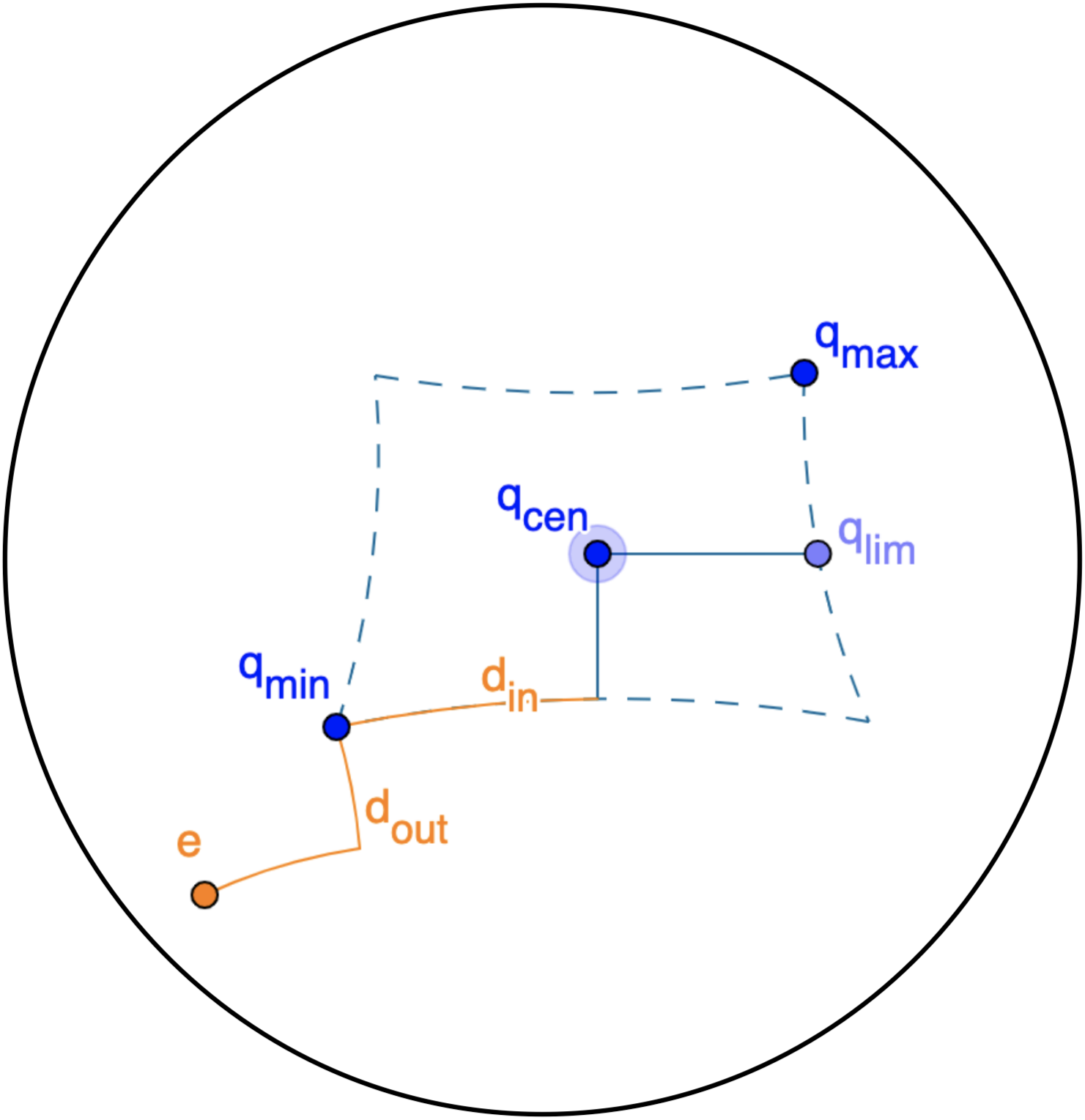}
		\caption{Hyperbolic distance}
		\label{fig:hyperbolic_distance}
	\end{subfigure}
		\caption{PFOE queries and hyperbolic distance in a Poincaré geodisc.}
		\Description[four figures]{The three figures show the three operations translation, intersection and union of hyperboloids. The fourth figure visualizes the hyperbolic distance.}
		\label{fig:three graphs}
\end{figure}

	\vspace{0.04in} \noindent \textbf{Translation (t):} Each relation $r \in R$ is equipped with a relation embedding $r = Hyperboloid_r \in \mathbb{R}^{2d}$. Given an entity embedding $e \in E$, we model its translation ($o_t$) and distance from the result entities $v_t \in V_t \subseteq E$ ($d_t$) as follows:
	\begin{equation}
		o_t = e \oplus_c r, \quad d_t(v) = d_{hyp}(v_t,o_t)\label{eq:translation}
	\end{equation} 
	This provides us with the translated hyperboloid with a new center and larger limit $(\text{lim}(r) \geq 0)$. A sample operation is illustrated in Figure \ref{fig:three graphs}\subref{fig:translation}.
	
	\vspace{0.04in} \noindent \textbf{Intersection ($\cap$):} We model the intersection of a set of hyperboloid embeddings $Q_\cap = \{e_1,e_2,e_3, ..., e_n\}$ as $o_{\cap}$  and entity distance from the result entities $v_\cap \in V_\cap \subseteq E$ as $d_\cap$ where:
	\begin{align}
		o_{\cap}&=(\text{cen}(Q_{\cap}), \text{lim}(Q_{\cap})) \label{eq:intersection}\\
		\text{cen}(Q_{\cap}) &= \sum_ia_i\odot_c \text{cen}(e_i); \quad a_i = \frac{\exp(f(e_i))}{\sum_j \exp(f(e_i)}\label{eq:att}\\
		\text{lim}(Q_{\cap}) &= \min(\{\text{lim}(e_1),...,\text{lim}(e_n)\}) \odot_c \sigma(DS(\{e_1,...,e_n\}))\nonumber\\
		DS(\{e_1,..,e_n\}) &= f\left(\frac{1}{n}\sum_{i=1}^n f(e_i)\right)\label{eq:ds}\\
		d_\cap(v_\cap) &= d_{hyp}(v_\cap,o_\cap)
	\end{align}
	where $\odot_c$ is the Möbius 
	scalar product, $f(.):\mathbb{R}^{2d}\rightarrow\mathbb{R}^d$ is the multilayer perceptron (MLP), $\sigma(.)$ is the sigmoid function and $DS(.)$ is the permutation invariant deep architecture, namely, DeepSets \cite{zaheer2017deep}. 
	$Min(.)$ and $\exp(.)$ are element-wise minimum and exponential functions.  The new center and limit are calculated by an attention layer \cite{bahdanau2014neural} over the hyperboloid centers and DeepSets for shrinking the limits, respectively. Figure \ref{fig:three graphs}\subref{fig:intersection} depicts a sample intersection.
	
	\vspace{0.04in} \noindent \textbf{Union ($\cup$): } Unlike intersection, union operations are not closed under hyperboloids (union of hyperboloids is not a hyperboloid).  Hence, the distance of entities from the union query space ($d_\cup$) is defined as the minimum distance from any hyperboloid in the union. For a set of hyperboloid embeddings $Q_\cup = \{e_1,e_2,e_3, ..., e_n\}$, union space is given by $o_{\cup}$  and distance from result entities $v_\cup \in V_\cup \subseteq E$ by $d_\cup$, where
	\begin{gather}
	o_{\cup}=Q_\cup\label{eq:union}\\
	d_\cup(v_\cup) = \min\left(\{d_{hyp}(v_\cup,e_i)~\forall e_i \in o_\cup\}\right)
	\end{gather}
	Note that, since union is not closed under hyperboloids it cannot be applied before the other operations. We circumvent this problem by utilizing Disjunctive Normal Form (DNF) transformation \cite{ren2020query2box} on our logical queries. This allows us to push all the union operations to the end of our computational graph, thus maintaining validity for all PFOE queries. An outline of HypE's training procedure is given in Algorithm \ref{alg:HypE}. 

\begin{algorithm}[htbp]
	\SetAlgoLined
	\KwIn{Training data $D_t, D_\cap, D_\cup$, which are set of all (query ($Q$), result ($V$)) for translation, intersection, and union, respectively;}
	\KwOut{ Entity $E$ and Relation $R$ hyperboloids; }
	Randomly initialize $e \in E$ and $r \in R$ ($e,r \in \mathbb{H}^{2d}$);\\
	\For{number of epochs; until convergence}
	{
	 $l = 0$;{\color{blue}{ \# Initialize loss}\\}
	\For{$\{(e,r, V_t) \in D_t\}$} 
	{
		\label{alg:start_translation}
		$o_t= e \oplus_c r$, from Eq. \eqref{eq:translation}\\
		\nonl {\color{blue}{ \# Update loss for translation queries}\\}
		\oldnl$l = l + \sum_{v_t \in V_t}d_{hyp}(v_t,o_t)$
	}\label{alg:end_translation}
	\For{$\{(Q_\cap, V_\cap) \in D_\cap\}$} 
	{	
		\label{alg:start_intersection}
		$o_\cap= (cen(Q_\cap),lim(Q_\cap))$, from Eq. \eqref{eq:intersection}\\
		\nonl {\color{blue}{ \# Update loss for intersection queries}\\}
		\oldnl$l = l + \sum_{v_\cap \in V_\cap}d_{hyp}(v_\cap,o_\cap)$
	} \label{alg:end_intersection}
	\For{$\{(Q_\cup, V_\cup) \in D_\cup\}$}
	{	
		 \label{alg:start_union}
		$o_\cup= Q_\cup$, from Eq. \eqref{eq:union}\\
		 \nonl{\color{blue}{ \# Update loss for union queries}\\}
		\oldnl$l = l + \sum_{v_\cup \in V_\cup}Min\left(d_{hyp}(v_\cup,e_i) ~\forall e_i \in o_\cup\right)$
	} \label{alg:end_union}
	{
		\nonl{\color{blue}{\# Update E and R with backpropagation}\\}
		\oldnl$E \leftarrow E - \Delta_E l$\\
		\oldnl$R \leftarrow R - \Delta_R l$
	}
	}
	\Return{E,R}
	\caption{HypE algorithm}
	\label{alg:HypE}
\end{algorithm}
\subsection{Implementation Details}
\label{sec:implementation}
{We implemented HypE in Pytorch \cite{NEURIPS2019_9015} on two Nvidia V100 GPUs with 16 GB VRAM. For gradient descent, the model is trained using Reimannian Adam optimizer \cite{becigneul2018riemannian} with an initial learning rate of 0.0001 and standard $\beta$ values of 0.9 and 0.999. We utilize ReLU \cite{10.5555/3104322.3104425} as the activation function. Also, we randomly selected 128 negative samples per positive sample in the training phase to learn better discriminative features. For our empirical studies, we learned hyperboloid embeddings of $2\times400$ dimensions (d=400)}. Due to the {conditionality (i.e., \textit{if} conditions in Algorithm \ref{alg:HypE})} in our computational graph, we employ a switch mechanism between the network layers \cite{DBLP:conf/ijcai/FangMZZB17,looks2017deep}. 
 The switch mechanism receives an operator signal that defines the operation and accordingly connects/disconnects a layer from the framework. 
 A disconnected switch blocks back-propagation of weight updates to the disconnected layers. This enables a concurrent use of all PFOE queries to update the entity and relation embeddings. For an input query $Q$ and resultant entities $V$, Algorithm \ref{alg:HypE} provides the pseudocode of our overall framework to learn representations of entities $E$ and relation $R$. The algorithm describes the three main operations, namely, translation (lines \ref{alg:start_translation}-\ref{alg:end_translation}), intersection (lines \ref{alg:start_intersection}-\ref{alg:end_intersection}), and union (lines \ref{alg:start_union}-\ref{alg:end_union}) \footnote{Implementation code: \url{https://github.com/amazon-research/hyperbolic-embeddings}}.
\section{Experimental Setup}
\label{sec:experimental}
This section describes the experimental setup that analyzes the performance of HypE on various problems. We aim to study the following research questions:

\begin{itemize}[leftmargin=0.5em]
	\item \textbf{RQ1:} For the task of reasoning over KGs, are hyperboloid embeddings better than the baselines at learning hierarchical relations?
	\item \textbf{RQ2:} What is the contribution of individual {components} in the HypE model?
	\item \textbf{RQ3:} Do the representations capture relevant {data features} for the downstream task of anomaly detection? 
	\item \textbf{RQ4:} Can hyperboloid embeddings leverage auxiliary semantic information from the entities?
	\item \textbf{RQ5:} Can we comprehend the latent representational space obtained by the proposed HypE model?
\end{itemize}

\subsection{Datasets}
\label{sec:dataset}
We perform our experimental study on the following standard KG and hierarchical graph datasets:
\begin{enumerate}
	\item \textbf{FB15k} \cite{NIPS2013_5071} contains knowledge base relation triples and textual mentions of Freebase entity pairs. This dataset contains a large number of simple test triples that can be obtained by inverting the training triples.
	\item \textbf{FB15k-237} \cite{toutanova-etal-2015-representing} is a subset of FB15k where all the simple inversible relations are removed, so the models can learn and focus on more complex relations.
	\item \textbf{NELL995} \cite{carlson2010toward} is a KG dataset of relation triples constructed from the $995^{th}$ iteration of the Never-Ending Language Learning (NELL) system.
	\item \textbf{DBPedia Hierarchical Taxonomy}\footnote{\url{https://www.kaggle.com/danofer/dbpedia-classes}} is a subset extracted from Wikipedia {snapshot} that provides multi-level hierarchical taxonomy over 342,782 articles (leaf-nodes). 
	\item \textbf{E-commerce Product Network}\footnote{Proprietary dataset} is a subsampled product taxonomy from an e-commerce platform. 
\end{enumerate}
To be consistent with KG terms, for the hierarchical graph datasets (DBPedia and E-commerce) we consider all the intermediate and leaf nodes as entities and the edges between them as relations.  Additionally, we consider two variants for encoding edges. First, all edges are considered identical ($|R|=1$) and second, where all edges are depth-encoded ($|R|=p$), where $p$ is maximum depth of the hierarchy ($p=3$  for DBPedia and $p=4$ for E-commerce dataset). 
For cross-validation and evaluation, we split the graph $KG$ into three parts: $KG_{train}$, $KG_{valid}$ and $KG_{test}$ in a $75:10:15$ ratio for our experiments. More details of the datasets are given in Table \ref{tab:dataset}.
\setlength{\tabcolsep}{6pt}
\begin{table}[htbp]
		\caption{Basic statistics of the datasets including the number of unique entities, relations, and edges.}
	\begin{tabular}{|c|c|c|c|c|c|}
		\hline
		Dataset&\# entities&\# relations&\# edges\\
		\hline
		FB15k&14,951&2690&273,710\\
		FB15k-237&14,505&474&149,689\\
		NELL995&63,361&400&107,982\\
		DBPedia&34,575&3&240,942\\
		E-commerce& $\sim$118K &4& $\sim$562K\\
		\hline
	\end{tabular}
	\vspace{-1em}
	\label{tab:dataset}
\end{table} 
\setlength{\tabcolsep}{3pt}
\subsection{Baselines}
\label{sec:baselines}
We select our baselines based on the following two criteria:
\begin{enumerate}
	\item The embedding geometries are closed under the intersection and translation operation{, e.g., the translation or intersection of arc-aligned hyperboloids results in an arc-aligned hyperboloid.}
	\item The baseline can be intuitively extended to all PFOE queries over KG. {This is necessary to have a fair comparison with HypE that can leverage all PFOE queries.}
\end{enumerate}
We adopt the following state-of-the-art baselines based on geometric diversity and our criterion to compare against HypE:
\begin{itemize}
	\item \textbf{Graph Query Embedding (GQE)} \cite{hamilton2018embedding} embeds entities and relations as a vector embedding in the Euclidean space. 
	\item \textbf{Knowledge Graph Attention Network (KGAT)} \cite{wang2019kgat} embeds entities and relations as a vector embedding in the Euclidean space utilizing attention networks over entities and relations with a TransR loss \cite{lin2017learning} . 
	\item \textbf{Hyperbolic Query Embeddings (HQE)} \cite{ganea2018hyperbolic} utilizes manifold transformations (refer to Section \ref{sec:poincare}) to represent entities and relations as a vector embedding in hyperbolic space. 
	\item \textbf{Query2Box (Q2B)} \cite{ren2020query2box} embeds entities and relations as axis-aligned hyper-rectangle or box embeddings in Euclidean space.
\end{itemize}
Some of the other possible baselines \cite{NIPS2013_5071,nickel2011three,yang2014embedding}, solely, focus on the multi-hop (or translation) problem. They could not be naturally extended to other PFOE queries. Additionally, other geometric variants such as circular and Gaussian embeddings \cite{pei2019dynamic,vilnis2014word} are not closed under intersection (intersection of Gaussians is not a Gaussian). 

\begin{table*}[htbp]
	\centering
	\footnotesize	
	\caption{Performance comparison of HypE (ours) against the baselines to study the {efficacy} of the Query-Search {space}. The columns present the different query structures and averages over them. The final row presents the Average Relative Improvement (\%) of HypE compared to Query2Box over all datasets. E-Vector and H-Vector are vectors in Euclidean and hyperbolic space, respectively. Best results for each dataset are shown in bold.}
	\begin{tabular}{|p{3.8em}|p{8.1em}|ccc|ccc|ccc|c|ccc|ccc|ccc|c|}
		\hline
		\textbf{Metrics}&&\multicolumn{10}{c|}{\textbf{Hits@3}}&\multicolumn{10}{c|}{\textbf{Mean Reciprocal Rank}}\\\hline
		\textbf{Dataset}&\textbf{Model}&\textbf{1t}&\textbf{2t}&\textbf{3t}&\boldmath$2\cap$&\boldmath$3\cap$&\boldmath$2\cup$&\boldmath$\cap t$&\boldmath$t\cap$&\boldmath$\cup t$&\textbf{Avg}&\textbf{1t}&\textbf{2t}&\textbf{3t}&\boldmath$2\cap$&\boldmath$3\cap$&\boldmath$2\cup$&\boldmath$\cap t$&\boldmath$t\cap$&\boldmath$\cup t$&\textbf{Avg}\\\hline
		FB15k&GQE (E-Vector)&.636&.345&.248&.515&.624&.376&.151&.310&.273&.386&.505&.320&.218&.439&.536&.300&.139&.272&.244&.330\\
&KGAT (E-Vector)&.711&.379&.276&.553&.667&.492&.181&.354&.302&.435&.565&.352&.243&.471&.573&.393&.167&.311&.270&.372\\
&HQE (H-Vector)&.683&.365&.265&.451&.589&.438&.135&.283&.290&.389&.543&.339&.233&.384&.506&.350&.125&.249&.259&.332\\
&Q2B (Box)&.786&.413&.303&.590&.710&.608&\textbf{.211}&.397&.330&.483&.654&.373&.274&.488&.602&.468&\textbf{.194}&.339&.301&.410\\
&HypE (Hyperboloid)&\textbf{.809}&\textbf{.486}&\textbf{.365}&\textbf{.598}&\textbf{.728}&\textbf{.610}&.206&\textbf{.406}&\textbf{.410}&\textbf{.513}&\textbf{.673}&\textbf{.439}&\textbf{.330}&\textbf{.495}&\textbf{.617}&\textbf{.470}&.189&\textbf{.347}&\textbf{.374}&\textbf{.437}\\\hline
		FB15k&GQE (E-Vector)&.404&.214&.147&.262&.390&.164&.087&.162&.155&.221&.346&.193&.145&.250&.355&.145&.086&.156&.151&.203\\
		-237&KGAT (E-Vector)&.436&.227&.167&.293&.422&.202&.069&.135&.174&.236&.373&.205&.165&.280&.384&.179&.068&.130&.170&.217\\
&HQE (H-Vector)&.440&.231&.171&.265&.387&.195&.083&.162&.183&.235&.376&.209&.169&.253&.352&.173&.082&.156&.179&.217\\
&Q2B (Box)&.467&.240&.186&.324&.453&\textbf{.239}&.050&.108&.193&.251&.400&.225&.173&.275&.378&\textbf{.198}&.105&.180&.178&.235\\
&HypE (Hyperboloid)&\textbf{.572}&\textbf{.366}&\textbf{.255}&\textbf{.399}&\textbf{.527}&.225&\textbf{.145}&\textbf{.246}&\textbf{.282}&\textbf{.335}&\textbf{.490}&\textbf{.343}&\textbf{.237}&\textbf{.339}&\textbf{.440}&.186&\textbf{.305}&\textbf{.410}&\textbf{.260}&\textbf{.334}\\\hline
		NELL&GQE (E-Vector)&.417&.231&.203&.318&.454&.200&.081&.188&.139&.248&.311&.193&.175&.273&.399&.159&.078&.168&.130&.210\\
		995&KGAT (E-Vector)&.486&.249&.218&.331&.467&.285&.107&.200&.151&.277&.362&.208&.188&.284&.410&.227&.103&.179&.141&.234\\
&HQE (H-Vector)&.477&.250&.219&.270&.413&.267&.091&.153&.166&.256&.355&.209&.189&.232&.363&.213&.088&.137&.155&.216\\
&Q2B (Box)&.555&.266&.233&.343&.480&.369&.132&.212&.163&.306&.413&.227&.208&.288&.414&.266&.125&.193&.155&.254\\
&HypE (Hyperboloid)&\textbf{.618}&\textbf{.359}&\textbf{.312}&\textbf{.400}&\textbf{.563}&\textbf{.441}&\textbf{.143}&\textbf{.227}&\textbf{.278}&\textbf{.371}&\textbf{.460}&\textbf{.306}&\textbf{.279}&\textbf{.336}&\textbf{.486}&\textbf{.318}&\textbf{.135}&\textbf{.207}&\textbf{.264}&\textbf{.310}\\\hline
		DBPedia&GQE (E-Vector)&.673&.006&N.A.&.873&.879&.402&.160&.668&0.00&.406&.502&.005&N.A.&.749&.773&.32&.154&.597&0.00&.344\\
		$|R|=1$&KGAT (E-Vector)&.753&.007&N.A.&.937&.940&.526&.192&.762&0.00&.457&.561&.006&N.A.&.804&.825&.419&.185&.682&0.00&.387\\
&HQE (H-Vector)&.422&.003&N.A.&\textbf{1.00}&\textbf{1.00}&.138&.109&.182&\textbf{.001}&.296&.314&.003&N.A.&\textbf{.859}&\textbf{.879}&.110&.105&.163&\textbf{.001}&.270\\
&Q2B (Box)&.832&.007&N.A.&\textbf{1.00}&\textbf{1.00}&.649&.224&.856&0.00&.508&.619&.006&N.A.&.840&.863&.468&.212&.779&0.00&.421\\
&HypE (Hyperboloid)&\textbf{.897}&\textbf{.009}&N.A.&\textbf{1.00}&\textbf{1.00}&\textbf{.708}&\textbf{.294}&\textbf{.935}&\textbf{.001}&\textbf{.546}&\textbf{.668}&\textbf{.008}&N.A.&.840&.863&\textbf{.511}&\textbf{.278}&\textbf{.853}&\textbf{.001}&\textbf{.447}\\\hline
		DBPedia&GQE (E-Vector)&.730&.565&N.A.&.873&.879&.534&.213&.705&.027&.504&.544&.421&N.A.&.651&.656&.398&.159&.526&.020&.375\\
		$|R|=p$&KGAT (E-Vector)&.816&.621&N.A.&.937&.940&.699&.255&.804&.030&.567&.608&.463&N.A.&.698&.700&.521&.190&.599&.022&.422\\
&HQE (H-Vector)&.456&.182&N.A.&\textbf{1.00}&\textbf{1.00}&.184&.143&.192&\textbf{.143}&.367&.339&.135&N.A.&\textbf{.744}&\textbf{.744}&.137&.106&.143&\textbf{.106}&.273\\
&Q2B (Box)&.901&.676&N.A.&\textbf{1.00}&\textbf{1.00}&.863&.297&.903&.033&.630&.670&.503&N.A.&\textbf{.744}&\textbf{.744}&.642&.221&.672&.025&.469\\
&HypE (Hyperboloid)&\textbf{.970}&\textbf{.756}&N.A.&\textbf{1.00}&\textbf{1.00}&\textbf{.940}&\textbf{.388}&\textbf{.985}&.046&\textbf{.676}&\textbf{.722}&\textbf{.563}&N.A.&\textbf{.744}&\textbf{.744}&\textbf{.700}&\textbf{.289}&\textbf{.733}&.034&\textbf{.503}\\\hline
		\multicolumn{2}{|c|}{\textbf{Avg. Improv. (\%) (HypE vs Q2B)}}&15.0&39.2&67.0&4.1&4.7&13.0&28.5&12.9&7.3&14.4&14.9&38.4&59.8&4.4&5.0&12.9&39.5&17.9&64.9&18.5\\\hline
	\end{tabular}
	\label{tab:efficiency}
		\vspace{-1.5em}
\end{table*}
\subsection{RQ1: {Efficacy} of the Query-Search space}
\label{sec:query}
\begin{figure}[htbp]
	\includegraphics[width=.88\linewidth]{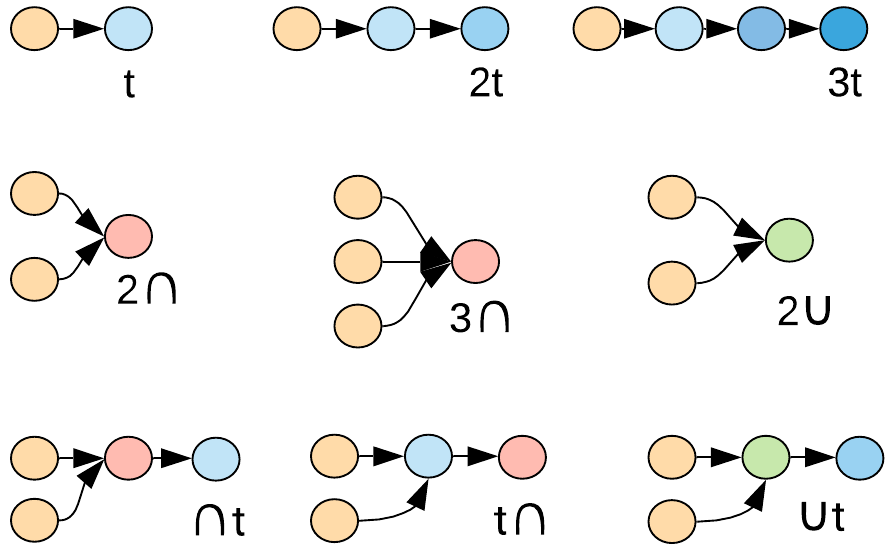}
	\caption{Logical {query structures} designed to compare HypE against baselines. The {\color{cyan}{blue}}, {\color{red}{red}}, and {\color{green}{green}} units denote translation, intersection, and union operations, respectively.}
	\Description[logical queries]{The figure shows the logical queries; both simple and compound queries.}
	\label{fig:queries}
\end{figure}

To analyze the efficacy of the query space obtained from the HypE model, we compare it against the state-of-the-art baselines on the following reasoning query structures:
\begin{enumerate}
	\item \textbf{Single operator queries}  include multi-level translation (1t, 2t, and 3t) multi-entity intersection ($2\cap, 3\cap$) and multi-entity union queries ($2\cup$). {1t, 2t, and 3t denote translation with 1, 2 and 3 consecutive relations, respectively. $2\cap$ and $2\cup$ stand for intersection and union over two entities, respectively. $3\cap$ represents intersection over three entities.}
	\item \textbf{Compound queries} contain multiple operators chained in series to get the final result. Our experiments analyze $\cap t$ (intersection-translation), $t \cap$ (translation-intersection) and $\cup t$ (union-translation).
\end{enumerate}
The above queries are illustrated in Figure \ref{fig:queries}. 

We extract the ground truth query-entity pairs by traversing the datasets \cite{ren2020query2box}. 
{The models are trained on queries from $KG_{train}$ and validated on $KG_{valid}$. The final evaluation metrics are calculated on $KG_{test}$.} {We utilize Euclidean norm and hyperbolic distance (given in Eq. \eqref{eq:hyp_dist}) to measure the distance between query embeddings and its resultant entities in Euclidean and hyperbolic spaces, respectively.} 
The sorted query-entity distances are the ranked results for the given query. 

\begin{table}[htbp]
	\centering
	\footnotesize	
	\caption{Performance comparison of HypE (ours) against the baselines on an e-commerce dataset for multi-level translation queries. We assume GQE (E-vector) as a baseline and report relative improvements against that for all the methods. The numbers are in percentages.  Best results for each dataset are shown in bold.\protect\footnotemark}
	\begin{tabular}{|l|l|ccc|ccc|}
		\hline
		\textbf{Metrics}&&\multicolumn{3}{c|}{\textbf{Hits@3}}&\multicolumn{3}{c|}{\textbf{Mean Reciprocal Rank}}\\\hline
		\textbf{Dataset}&\textbf{Model}&\textbf{1t}&\textbf{2t}&\textbf{3t}&\textbf{1t}&\textbf{2t}&\textbf{3t}\\\hline
		$|R|=1$&GQE (E-Vector)&0.0&0.0&0.0&0.0&0.0&0.0\\
		&KGAT (E-Vector)&12.12&10.9&0.0&0.0&-36.2&-57.3\\
		&HQE (H-Vector)&-34.8&-51.6&0.0&-34.7&-56.6&0.0\\
		&Q2B (Box)&30.3&4.7&\textbf{200}&23.5&7.5&0.0\\
		&HypE (Hyperboloid)&\textbf{38.6}&\textbf{98.4}&\textbf{200}&\textbf{38.8}&\textbf{79.2}&\textbf{100}\\\hline
		$|R|=p$&GQE (E-Vector)&0.0&0.0&0.0&0.0&0.0&0.0\\
		&KGAT (E-Vector)&12.2&10.7&0.0&11.5&9.5&0.0\\
		&HQE (H-Vector)&38.8&92.9&350&38.5&90.5&600\\
		&Q2B (Box)&17.3&37.5&0.0&23.1&19.0&0.0\\
		&HypE (Hyperboloid)&\textbf{122.3}&\textbf{298.2}&\textbf{8350}&\textbf{192.3}&\textbf{471.4}&\textbf{9900}\\\hline
	\end{tabular}
	\label{tab:efficiencyecomm}
	\vspace{-1.5em}
\end{table}
\begin{table*}[htbp]
	\centering
	\footnotesize
	\caption{Ablation study results. {The first column presents the model variants compared against the final HypE model. Avg-1t and Avg-1,2,3t variants only utilize average center aggregation because other aggregation variants only apply when intersections are involved. HypE-TC presents the HypE variant with trainable curvature. The metrics reported in the table are averaged across evaluation on all the datasets. Finer evaluation results are provided in the Appendix \ref{app:ablation}. Best results are shown in bold.}}
	\begin{tabular}{|l|ccc|ccc|ccc|c|ccc|ccc|ccc|c|}
		\hline
		\textbf{Metrics}&\multicolumn{10}{c|}{\textbf{Hits@3}}&\multicolumn{10}{c|}{\textbf{Mean Reciprocal Rank}}\\\hline
		\textbf{Model Variants}&\textbf{1t}&\textbf{2t}&\textbf{3t}&\boldmath$2\cap$&\boldmath$3\cap$&\boldmath$2\cup$&\boldmath$\cap t$&\boldmath$t\cap$&\boldmath$\cup t$&\textbf{Avg}&\textbf{1t}&\textbf{2t}&\textbf{3t}&\boldmath$2\cap$&\boldmath$3\cap$&\boldmath$2\cup$&\boldmath$\cap t$&\boldmath$t\cap$&\boldmath$\cup t$&\textbf{Avg}\\\hline
		HypE-Avg-1t&.651&.274&.153&.383&.414&.484&.099&.235&.135&.312&\textbf{.538}&.238&.149&.309&.332&.366&.108&.216&.126&.264\\
		HypE-Avg-1,2,3t&.650&.353&.213&.431&.474&.486&.133&.280&.187&.352&.532&.306&.204&.345&.381&.366&.140&.256&.171&.297\\
		HypE-Avg&.650&.432&.272&.479&.535&.487&.167&.325&.239&.392&.525&.373&.258&.381&.429&.366&.172&.296&.215&.330\\
		HypE-DS&.630&.432&.267&.488&.555&.451&.165&.356&.238&.392&.511&.376&.257&\textbf{.399}&\textbf{.460}&.344&.172&.326&.213&.334\\
		HypE-TC&\textbf{.656}&\textbf{.439}&\textbf{.275}&\textbf{.490}&\textbf{.565}&\textbf{.488}&\textbf{.177}&\textbf{.374}&\textbf{.243}&\textbf{.406}&.530&\textbf{.379}&\textbf{.262}&.391&.458&\textbf{.368}&\textbf{.184}&\textbf{.340}&\textbf{.216}&\textbf{.342}\\\hline
		HypE (final)&\textbf{.656}&.438&\textbf{.275}&\textbf{.490}&.564&\textbf{.488}&\textbf{.177}&.373&.242&.405&.530&.378&\textbf{.262}&.390&.458&\textbf{.368}&\textbf{.184}&\textbf{.340}&.215&.341\\\hline
	\end{tabular}
	\label{tab:ablation}
	\vspace{-1.5em}
\end{table*}
\footnotetext{Due to the sensitive nature of the proprietary dataset, we only report relative results.}

Given {a} test query $q_{test}$, let the true ranked result entities be $E_{result}$ and model's ranked output be $E_{output} = \{e_{o1}, e_{o2},...,e_{on}\}$. The evaluation metrics {used in our work} are \textbf{Hits@K} and \textbf{Mean Reciprocal Rank (MRR)}. The metrics are {given by}:
\begin{align*}
HITS @K(q_{test}) &= \frac{1}{K}\sum_{k=1}^Kf\left(e_{ok}\right), ~ &f\left(e_{ok}\right) = \begin{cases}
1,\text{if }e_{ok} \in E_{result}\\
0,\text{ otherwise}
\end{cases}\\
MRR(q_{test})&= \frac{1}{n}\sum_{i=1}^n\frac{1}{f\left(e_{oi}\right)}, ~ &{f\left(e_{oi}\right)} = \begin{cases}
i, \text{if }e_{oi} \in E_{result}\\
\infty, \text{otherwise}
\end{cases} 
\end{align*} 

From the results in Table \ref{tab:efficiency}, we observe that, on average, HypE outperforms the current baselines in translation, single, and compound operator queries by 15\%-67\%, 4\%-13\%, and 7\%-28\%, respectively. The performance improvement linearly increases with higher query depth ($1t<2t<3t$). Furthermore, we notice that unique depth encoding ($|R|=d$) outperforms identical depth encoding ($|R|=1$) by 23\% in DBPedia. Because of our subsampling strategy, the E-commerce Product Network is disjoint (i.e., there are several intermediate nodes that do not share children). Hence, the number of intersection queries are extremely low and insufficient for training HypE or baselines. However, we can still train the translation queries, and the results are shown in Table \ref{tab:efficiencyecomm}, where we report relative performance improvements with respect to the GQE baseline. For the sake of completeness, results on intersection and union queries are given in the Appendix \ref{app:e-commerce}. 


\subsection{RQ2: Ablation Study}
\label{sec:ablation}
In this section, we empirically analyze the importance of different layers adopted in the HypE model. For this, we experiment with different variations of the center aggregation layer; Average (HypE-Avg), Attention (HypE) (refer to Eq. \eqref{eq:att})  and Deepsets (HypE-DS) (refer to Eq. \eqref{eq:ds}). Furthermore, we test the exclusion of intersection and unions to {comprehend} their importance in the representation learning process. We adopt two variants of HypE; one trained on only 1t queries (HypE-Avg-1t) and the other trained on all translation queries (HypE-Avg-1,2,3t). Table \ref{tab:ablation} presents the performance metrics of different variants on the query processing task, averaged across all the datasets including the E-commerce dataset. The results are provided in Table \ref{tab:ablation}. 

Firstly, we observe that the exclusion of intersection and union queries results in a significant performance decrease by 25\% (Avg-1,2,3t vs HypE). Furthermore, removing deeper queries such as 2t and 3t, also results in an additional decrease by 17\% (Avg-1t vs Ag-1,2,3t).  The tests on different aggregation layers prove that Attention is better than average and Deepsets by 23.5\% and 14.5\%, respectively. Additionally, we notice that employing a trainable curvature results in a slight performance improvement of 0.3\%. However, given the incremental performance boost but significant increase in the number of parameters ($\sim$10K) that the trainable curvature adds to the framework, {we ignore this component in the final HypE model.}

As explained in Section \ref{sec:reasoning}, the final HypE model adopts a Poincaré ball manifold with non-trainable curvature, in addition to attention and Deepsets layer for center and limit aggregation, respectively. Additionally, HypE leverages all PFOE queries.

\subsection{RQ3: Performance on Anomaly Detection}
\label{sec:anomaly}
In this experiment, we utilize the entity and relation representations, trained on the DBPedia Hierarchical Taxonomy and E-commerce Product Network with query processing task, to identify products that might be anomalously categorized. We consider identifying the anomalous children by three levels of parents (i.e., taxonomy levels); $P_1$, $P_2$ and $P_3$. The motivating application is to categorize items that are potentially mis-categorized by sellers into the more relevant (correct) part of the product taxonomy.

\begin{table}[htbp]
	\caption{Results on Miscategorized Article Anomaly Detection in DBPedia dataset. Best results are shown in bold and the second best results are underlined. P, R, and F1 represent Precision, Recall, and F-score, respectively.}
	\centering
	\footnotesize
	\begin{tabular}{|l|l|ccc|ccc|ccc|}
		\hline
		&\textbf{P-Level}&\multicolumn{3}{|c|}{\textbf{1}}&\multicolumn{3}{|c|}{\textbf{2}}&\multicolumn{3}{|c|}{\textbf{3}}\\\hline
		\textbf{Dataset}&\textbf{Models}&\textbf{P}&\textbf{R}&\textbf{F1}&\textbf{P}&\textbf{R}&\textbf{F1}&\textbf{P}&\textbf{R}&\textbf{F1}\\\hline
		DBPedia&GQE  &.512&.369&.428&.549&.446&.492&.576&.409&.479\\
		$|R|=p$&KGAT &.523&.375&.437&.552&.448&.495&.578&.416&.484\\
		&HQE  &.529&.385&.446&.556&.45&.497&.586&.424&.492\\
		&Q2B  &.589&\underline{.479}&.527&.589&.479&.528&.597&.481&.532\\\hline
		&HypE &.590&\underline{.479}&.528&\underline{.648}&.482&.552&.650&\underline{.486}&\underline{.557}\\
		&HypE-SI &\underline{.591}&\underline{.479}&\underline{.529}&\underline{.648}&.\underline{483}&\underline{.553}&\underline{.651}&\underline{.486}&\underline{.557}\\
		&HypE-SC&\textbf{.601}&\textbf{.501}&\textbf{.546}&\textbf{.662}&\textbf{.563}&\textbf{.608}&\textbf{.705}&\textbf{.563}&\textbf{.626}\\\hline
	\end{tabular}
	\label{tab:anomaly_semantics}
	\vspace{-1.5em}
\end{table}

\begin{table}[htbp]
	\caption{Results on Miscategorized Product Anomaly Detection in E-commerce Product Networks. Best results are shown in bold and the second best results are underlined. The improvements are relative to the GQE baseline. P, R, and F1 represent Precision, Recall, and F-score, respectively.\protect\footnotemark}
	\centering
	\footnotesize
	\begin{tabular}{|l|l|ccc|ccc|ccc|}
		\hline
		&\textbf{P-Level}&\multicolumn{3}{|c|}{\textbf{1}}&\multicolumn{3}{|c|}{\textbf{2}}&\multicolumn{3}{|c|}{\textbf{3}}\\\hline
		\textbf{Dataset}&\textbf{Models}&\textbf{P}&\textbf{R}&\textbf{F1}&\textbf{P}&\textbf{R}&\textbf{F1}&\textbf{P}&\textbf{R}&\textbf{F1}\\\hline
		E-&GQE&0.0&0.0&0.0&0.0&0.0&0.0&0.0&0.0&0.0\\
		commerce&KGAT &2.2&1.8&2.1&0.6&0.5&0.7&0.2&1.6&1.1\\
		$|R|=p$&HQE  &3.4&4.5&4.1&1.2&1.0&1.1&1.7&3.5&2.8\\
		&Q2B   &15.1&\underline{29.9}&23.4&7.2&7.4&7.4&3.6&17.5&11.3\\\hline
		&HypE &15.3&\underline{29.9}&23.4&\underline{18.0}&\underline{8.4}&\underline{12.8}&\underline{16.6}&18.5&\underline{17.2}\\
		&HypE-SI  &\underline{15.5}&\underline{29.9}&\underline{23.7}&\underline{18.0}&\underline{8.4}&12.5&16.4&\underline{18.8}&\underline{17.2}\\
		&HypE-SC&\textbf{17.4}&\textbf{35.8}&\textbf{27.5}&\textbf{20.6}&\textbf{26.4}&\textbf{23.7}&\textbf{22.3}&\textbf{37.6}&\textbf{30.8}\\\hline
	\end{tabular}
	\label{tab:anomaly_semantics_ecomm}
	\vspace{-1.5em}
\end{table}
\footnotetext{Due to the sensitive nature of the proprietary dataset, we only report relative results.}
\begin{table*}[htbp]
	\centering
	\footnotesize
	\caption{Example of Anomalies in the E-commerce dataset. The models predict ``\texttt{MISCAT}'' and ``\texttt{TRUE}'' tags for mis-categorized and truly-categorized items, respectively. Correct and Incorrect tags are given in {\color{green}green} and {\color{red} red} color, respectively. HypE performs better than Query2Box (Q2B) as we consider higher level of parents because hyperbolic space is better at capturing hierarchical features. Also, HypE-SC is able to utilize semantic information to improve prediction.}
	\begin{tabular}{p{40em}|cc|ccc}
		\hline
		\textbf{Product Title}&\multicolumn{2}{|c|}{\textbf{Parent}}&\multicolumn{3}{|c}{\textbf{Prediction}}\\
		&Category&P-level&Q2B&HypE&HypE-SC\\\hline
		ASICS Women's Gel-Venture 7 Trail Running Shoes, 5.5M, Graphite Grey/Dried Berry&Rompers&1&{\color{red}{TRUE}}&{\color{red}{TRUE}}&{\color{green}{MISCAT}}\\
		inktastic Family Cruise Youth T-Shirt Youth X-Large (18-20) Pacific Blue 35074&Calvin Klein&1&{\color{red}{TRUE}}&{\color{green}{MISCAT}}&{\color{green}{MISCAT}}\\
		Calvin Klein
		Men's Cotton Classics Multipack Crew Neck T-Shirts&Calvin Klein&1&{\color{red}{MISCAT}}&{\color{red}{MISCAT}}&{\color{green}{TRUE}}\\\hline
		Epic Threads Big Girls Paint Splatter Distressed Girlfriend Denim Jeans (Dark Wash, 10)&Wool \& Blends&2&{\color{red}{MISCAT}}&{\color{red}{MISCAT}}&{\color{green}{TRUE}}\\
		New Balance Women's Crag V1 Fresh Foam Trail Running Shoe, Black/Magnet/Raincloud&Wool \& Blends&2&{\color{red}{TRUE}}&{\color{green}{MISCAT}}&{\color{green}{MISCAT}}\\
		Fifth Harmony Vintage Photo Blue T Shirt (M)&Customer Segment&2&{\color{red}{TRUE}}&{\color{red}{TRUE}}&{\color{green}{MISCAT}}\\
		Billy Bills Playoff Shirt Buffalo T-Shirt Let’s Go Buffalo Tee&Customer Segment&2&{\color{red}{MISCAT}}&{\color{green}{TRUE}}&{\color{green}{TRUE}}\\
		\hline
		Kanu Surf Toddler Karlie Flounce Beach Sport 2-Piece Swimsuit, Ariel Blue, 4T&Brand Stores&3&{\color{red}{TRUE}}&{\color{green}{MISCAT}}&{\color{green}{MISCAT}}\\
		The North Face Infant Glacier ¼ Snap, Mr. Pink, 0-3 Months US Infant&Brand Stores&3&{\color{red}{MISCAT}}&{\color{green}{TRUE}}&{\color{green}{TRUE}}\\
		Artisan Outfitters Mens Surfboard Shortboard Batik Cotton Hawaiian Shirt&Specialty Stores&3&{\color{red}{MISCAT}}&{\color{green}{TRUE}}&{\color{green}{TRUE}}\\
		PUMA Unisex-Kid's Astro Kick Sneaker, Peacoat-White-teamgold, 3 M US Little Kid&Specialty Stores&3&{\color{red}{TRUE}}&{\color{red}{TRUE}}&{\color{green}{MISCAT}}\\
		\hline
	\end{tabular}
	\label{tab:amazon_examples}
		\vspace{-1.5em}
\end{table*}
\begin{figure*}[htbp]
	\centering
	\begin{subfigure}[b]{.49\linewidth}
		\centering
		\includegraphics[width=.8\linewidth]{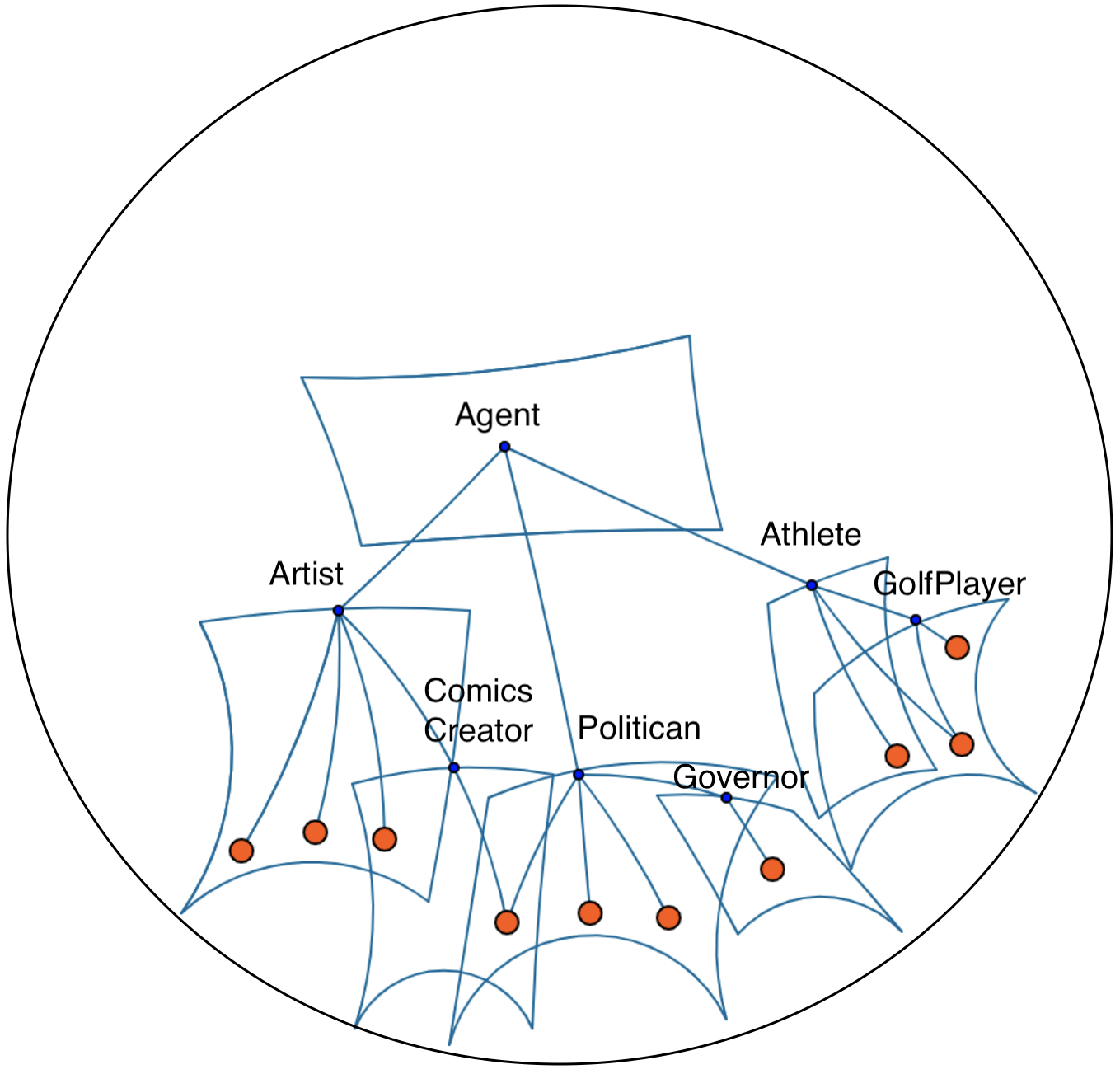}
		\caption{DBPedia  Taxonomy}
		\label{fig:dbpedia_taxonomy}
	\end{subfigure}
	\begin{subfigure}[b]{.49\linewidth}
		\centering
		\includegraphics[width=.8\linewidth]{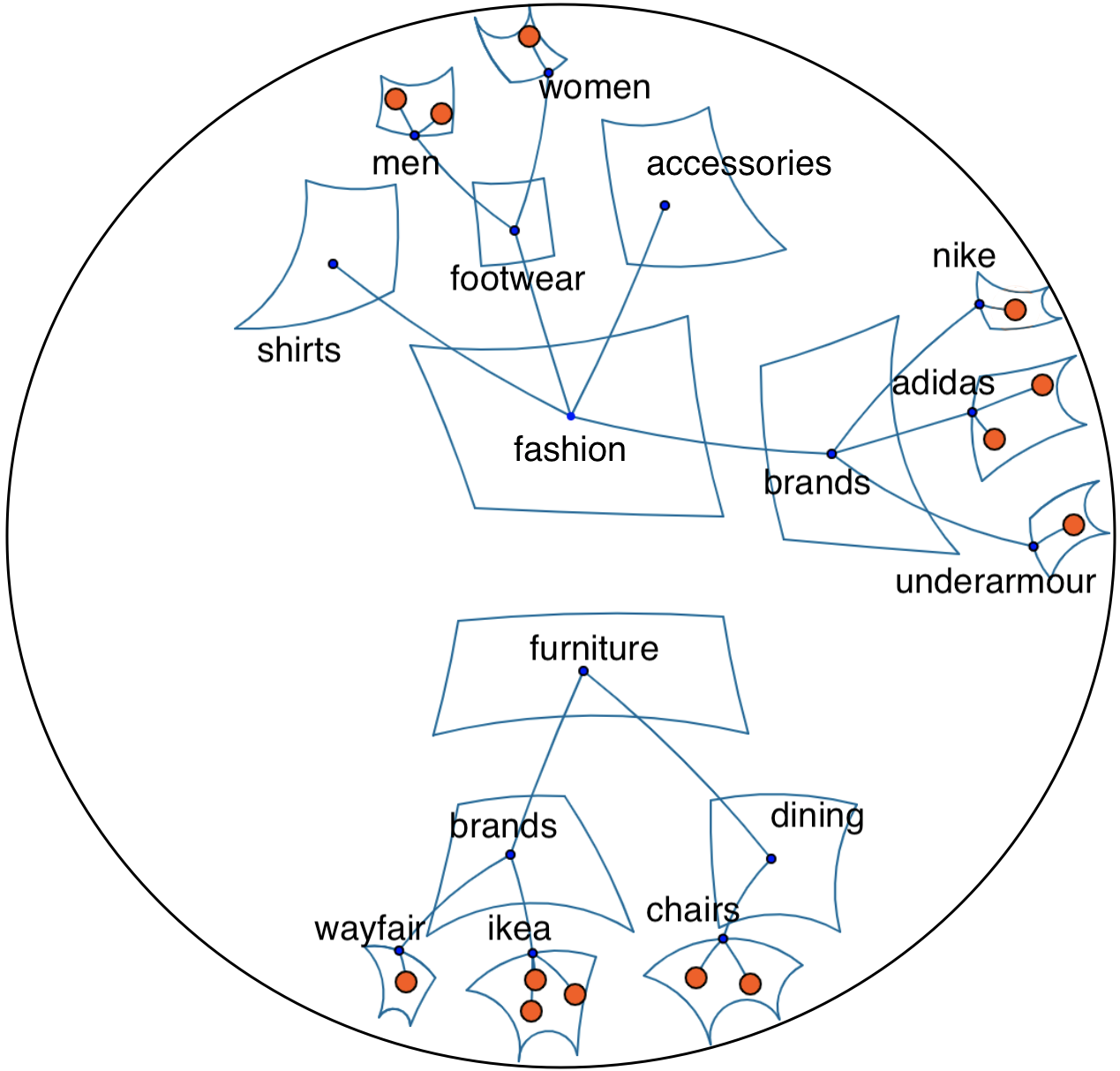}
		\caption{E-commerce Product Networks}
		\label{fig:amazon_product}
	\end{subfigure}
	\caption{Visualization of HypE representations for samples from hierarchical datasets  in Poincaré ball. The hyperboloids have been scaled up 10 times for better comprehension. The {\color{royalblue}{blue}} (intermediate nodes) circles are annotated with their entity names and {\color{darkorange}{orange}} circles (leaf nodes) depict articles and products in (a) and (b), respectively.}
	\Description[HypE visualization]{The figure shows our datasets interpretation in the hyperbolic space. The figure how higher levels of hierarchy generally contain more samples and hence have spatially larger hyperboloids.}
	\label{fig:visualization}
	\vspace{-1.5em}
\end{figure*}
We construct a pseudo-tree, where all parent nodes are infused with 10\% noise of randomly sampled anomalous leaf nodes from different parts of the dataset. The {goal} of the model is to learn representations from this pseudo-tree and {identify anomalous} leaf nodes of the {immediate} parent nodes. 
From the set of all intermediate nodes $KG_P$, given a parent $p \in KG_P$  and its originial set of children $C_{+}=children(p)$ and randomly sampled set of anomalous children $C_{-} = children\left( KG_P \setminus \{p\} \right)$, the aim here is to identify $c \in C_{-}$ from $C_{+} \cup C_{-}$. We use Precision, Recall and F1-score as the evaluation metrics for this experiment.

{The results (given in Table \ref{tab:anomaly_semantics} and Table \ref{tab:anomaly_semantics_ecomm}) show that, although HypE has comparable performance to baselines at $P_1$, it outperforms the baselines by more than $5 \%$ at $P_2$ and $P_3$. This demonstrates the robustness of HypE to noisy data and its capability of capturing relevant hierarchical features (as F1 on $P_3 > P_2 > P_1$) for downstream tasks}. Furthermore, the specific task is critical in e-commerce search {as irrelevant results impede a smooth user experience. Table \ref{tab:amazon_examples} presents some qualitative examples from the E-commerce dataset.} 

\subsection{RQ4: Leveraging Semantic Information}
\label{sec:semantic}
KGs generally also contain additional auxiliary information within the entities. In this section, we test the possibility of leveraging the semantic information in the DBPedia (article titles) and E-commerce (product titles) dataset to improve representations. 
We {study} two methods to connect HypE with FastText embeddings \cite{bojanowski2016enriching} of the corresponding titles:
\begin{itemize}[leftmargin=*]
	\item \textbf{Semantic Initiation (SI)} initiates the HypE's entities with semantic embeddings and learns new HypE-SI embeddings with the query-processing task (given in Section \ref{sec:query}).
	 \item \textbf{Semantic Collaboration (SC)} concatenates the HypE's pre-trained entity representations with semantic embeddings.
\end{itemize}
We investigate the performance of these methods on the task of anomaly detection. The results of the experiments are given in Tables \ref{tab:anomaly_semantics} and \ref{tab:anomaly_semantics_ecomm}. 
The results demonstrate that HypE-SI shows no significant performance improvement over HypE. That is, a good semantic initialization of the vectors does not result in better representations. This is reasonable, since the semantic embeddings are learnt in the Euclidean space, and several transformations occur between the initialization and final representations. This also means that the learning framework is robust to initialization. 
We observe a performance improvement of $2\%-8\%$ in case of HypE-SC when compared to HypE. This suggests the ubiquity of HypE since hierarchical representations can be independently augmented with other auxiliary features to solve more complex tasks. From the examples given in Table \ref{tab:amazon_examples}, we can observe that HypE-SI is able to leverage semantic information from \textit{product title} and \textit{category name} to enrich HypE's hierarchical information to produce better predictions. The additional semantic information is especially useful for product miscategorization. In the absence of semantic information, HypE will merely learn representations based on the noisy graph and will lose discriminative information between outliers and correct nodes.

\subsection{RQ5: Visualization of the Poincaré ball}
\label{sec:visualization}
We extract representative examples of different operations from our dataset and visualize them in a 2-dimensional Poincaré ball of unit radius. We employ Principal Component Analysis (PCA) \cite{bishop1995neural} for dimensionality reduction $(\mathbb{R}^{2d}\rightarrow\mathbb{R}^2)$ of the hyperboloids in Poincaré ball. 

Figure \ref{fig:visualization} depicts the HypE representations in a Poincaré ball manifold. Notice that the density of nodes {increases superlinearly} from the center towards the circumference, which is analogous to the {superlinear} increase in the number of nodes from root to the leaves. {Thus, HypE is able to learn a better distinction between different hierarchy levels, when compared to the Euclidean distance-based baselines, which conform to a linear increase.} Furthermore, we observe that hyperboloid intersections in DBPedia Taxonomy (Figure \ref{fig:dbpedia_taxonomy}) capture entities with common parents. Also, the disjoint nature of E-commerce Product Networks (Figure 
\ref{fig:amazon_product}) is illustrated by disjoint non-intersecting hyperboloids in the latent space. {In addition, we can also notice that the learnable limit parameter adjusts the size of hyperboloids to accommodate its varying number of leaf nodes. Thus, the complex geometry of HypE is able to improve its precision over vector baselines that, generally, utilize static thresholds over distance of the resultant entities from query points.}

\section{Conclusion}
\label{sec:conclusion}
In this paper, we presented Hyperboloid Embeddings (HypE) model, a novel self-supervised learning framework that utilizes dynamic query-reasoning over KGs as a proxy task to learn representations of entities and relations in a hyperbolic space. We demonstrate the efficacy of a hyperbolic query-search space against state-of-the-art baselines over different datasets. Furthermore, we also show the effectiveness of hyperboloid representations in complex downstream tasks and study methods that can leverage node's auxiliary information to enrich HypE features. Additionally, we analyze the contribution of HypE's individual components through an ablation study. Finally, we present our hyperboloid representations in a 2-dimensional Poincaré ball for better comprehensibility.
\bibliographystyle{ACM-Reference-Format}
\bibliography{sample-base}

\appendix
\section{Hyperbolic vs Euclidean distances}
\label{app:hyperbolic}
To better analyze the impact of adopting hyperbolic space, we need to understand its distinction from the Euclidean space in handling hierarchy. For this, we study the intra-level and inter-level Euclidean and hyperbolic distance between entities at different levels of the dataset. Let us say $E_p$ is the set of entities at level $P_p$ in the E-commerce Product Networks dataset. For the analysis, we calculate two sets of distances; intra-level ($\Delta_{intra}$) and inter-level ($\Delta_{inter}$) distance as follows:
	\begin{align}
	\Delta_{intra}(P_p) &= \frac{\sum_{i=0}^{|E_p|}\sum_{j=i}^{|E_p|}\delta(e_i,e_j)}{|E_p|\left(|E_p|-1\right)}\\ \Delta_{inter}(P_{p1},P_{p2}) &=\frac{\sum_{i=0}^{|E_{p1}|}\sum_{j=0}^{|E_{p2}|}\delta(e_i,e_j)}{|E_{p1}|\times|E_{p2}|}
	\end{align}
	$\delta(.~,~.)$ is replaced with Euclidean norm (on Query2Box representations) and hyperbolic distance ($d_{hyp}$ on HypE representations) to understand the difference between the hierarchical separation of entities in the two spaces. 
	
\begin{figure}[htbp]
	\includegraphics[width=.76\linewidth]{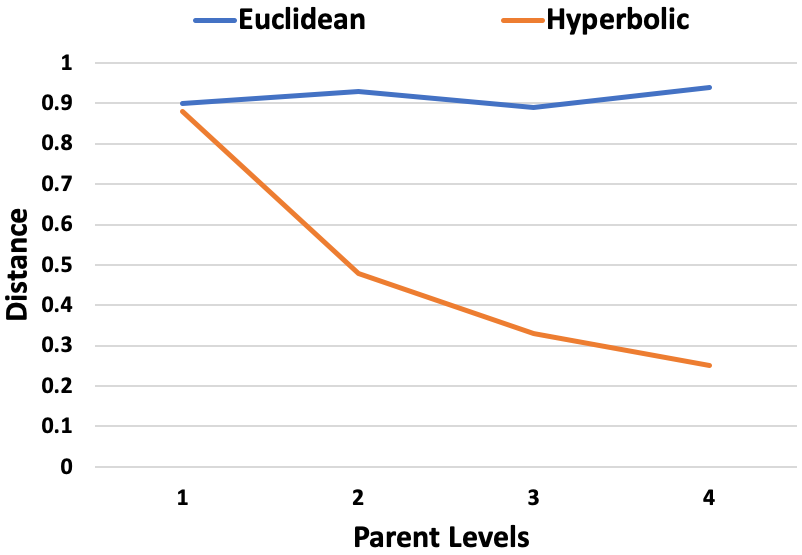}
	\caption{Intra-level \textcolor{royalblue}{Euclidean (Q2B)} and \textcolor{orange}{Hyperbolic (HypE)} distances. The graph presents $\Delta_{intra}$ of entity sets at different hierarchy levels, given on the x-axis.}
	\Description[Intra-level Distance functions]{Hyperbolic distance between intra-level entities is decreasing over levels, whereas, the Euclidean distance remains more or less constant.}
	\label{fig:intra_distance}
\end{figure}

\begin{figure}[htbp]
	\includegraphics[width=\linewidth]{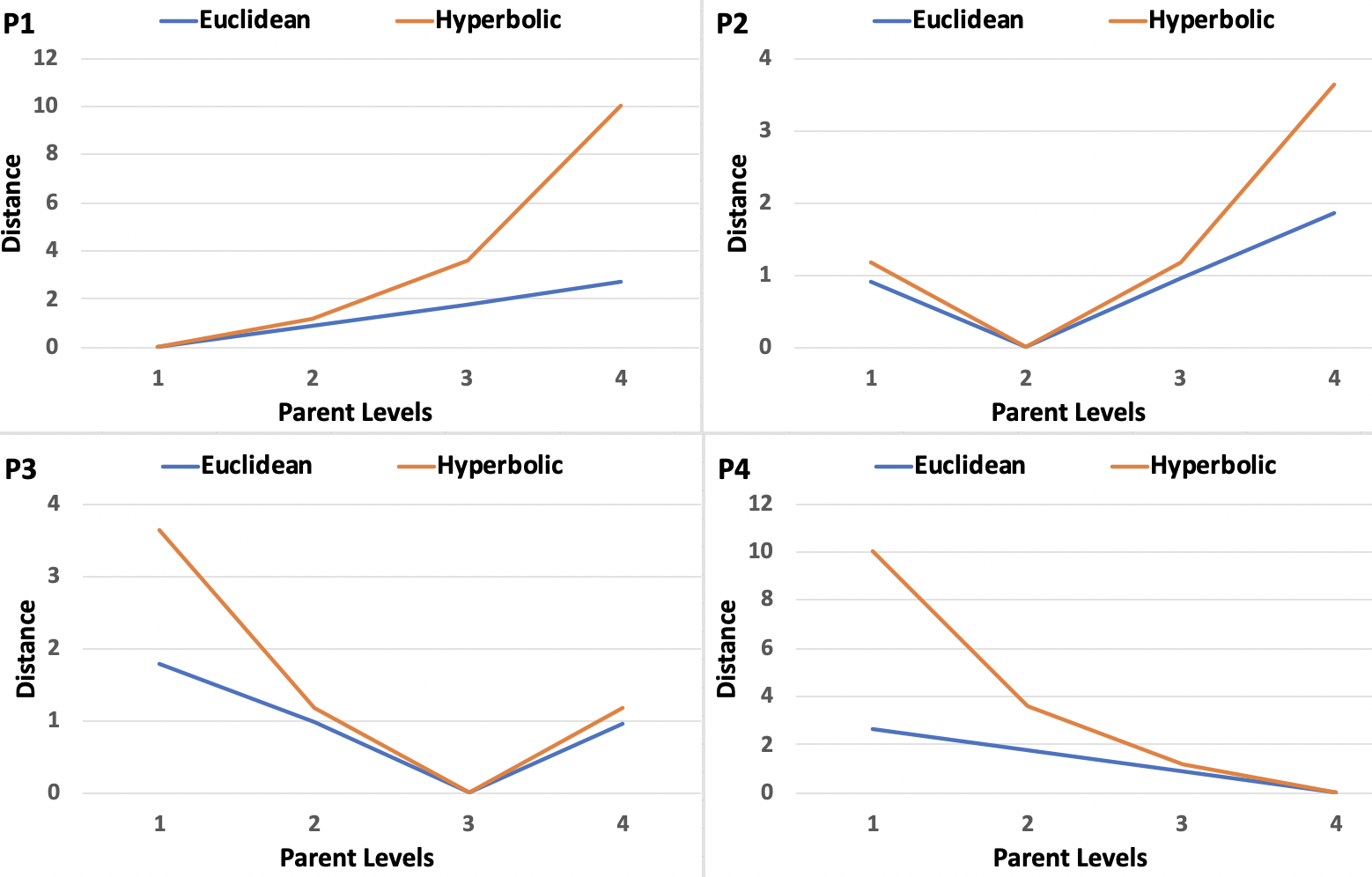}
	\caption{Inter-level \textcolor{royalblue}{Euclidean (Q2B)} and \textcolor{orange}{Hyperbolic (HypE)} distances. Each graph presents $\Delta_{inter}$ between entity pairs of a source hierarchical level, given by the graph label and the other hierarchy levels in the dataset, given on the x-axis.}
	\label{fig:inter_distance}
	\Description[Inter-level Distance functions]{Hyperbolic distance between inter-level entities is super-linear, compared to the linearly increasing distance of Eulcidean distances.}
\end{figure}

	In the $\Delta_{intra}$ results (depicted in Figure \ref{fig:intra_distance}), we observe that, with increasing level of hierarchy the distance between entities at different levels remains constant in the case of Euclidean space and shows a clear decreasing trend for hyperbolic space. This indicates denser clustering of entities at the same level. Additionally, the $\Delta_{inter}$ results, illustrated in Figure \ref{fig:inter_distance}, depict a linear increase in distance between inter-level entities in the Euclidean space and a superlinear growth in the hyperbolic space. This shows that hyperbolic space also learns clusters such that inter-level entities are farther apart compared to Euclidean space. This nature of inter-level discrimination and intra-level aggregation demonstrates the superior ability of hyperbolic spaces at capturing hierarchical features.
\begin{table*}[htbp]
	\centering
	\footnotesize	
	\caption{Performance comparison of HypE (ours) against its variants to study the {efficacy} of its Query-Search {space}. The columns present different query structures and averages values. Best results for each dataset are shown in bold.}
	\begin{tabular}{|p{3.6em}|p{8.1em}|ccc|ccc|ccc|c|ccc|ccc|ccc|c|}
		\hline
		\textbf{Metrics}&&\multicolumn{10}{c|}{\textbf{Hits@3}}&\multicolumn{10}{c|}{\textbf{Mean Reciprocal Rank}}\\\hline
		\textbf{Dataset}&\textbf{Model}&\textbf{1t}&\textbf{2t}&\textbf{3t}& \boldmath$2\cap$ & \boldmath$3\cap$ & \boldmath$2\cup$ & \boldmath$\cap t$ & \boldmath$t\cap$ & \boldmath$\cup t$ & \textbf{Avg}&\textbf{1t}&\textbf{2t}&\textbf{3t}& \boldmath$2\cap$ & \boldmath$3\cap$ &\boldmath$2\cup$&\boldmath$\cap t$&\boldmath$t\cap$&\boldmath$\cup t$&\textbf{Avg}\\\hline
		FB15k&HypE-Avg-1t&.803&.304&.202&.468&.534&.606&.115&.256&.228&.395&\textbf{.683}&.276&.188&.392&.448&.467&.111&.221&.219&.338\\
&HypE-Avg-1,2,3t&.803&.392&.282&.527&.612&.608&.155&.305&.317&.446&.675&.355&.257&.438&.513&.467&.144&.262&.296&.380\\
&HypE-Avg&.802&.479&.361&.585&.690&.609&.194&.353&.405&.497&.667&.433&.325&.483&.578&.467&.177&.302&.373&.422\\
&HypE-DS&.777&.479&.354&.596&.717&.564&.192&.387&.403&.496&.649&.437&.324&\textbf{.506}&\textbf{.620}&.439&.176&.333&.370&.428\\
&HypE-TC&\textbf{.809}&\textbf{.486}&\textbf{.365}&\textbf{.598}&\textbf{.729}&\textbf{.610}&\textbf{.206}&\textbf{.406}&\textbf{.410}&\textbf{.513}&.674&\textbf{.439}&\textbf{.331}&.495&.617&\textbf{.470}&\textbf{.189}&\textbf{.348}&\textbf{.375}&\textbf{.437}\\
&HypE (final)&\textbf{.809}&\textbf{.486}&\textbf{.365}&\textbf{.598}&.728&\textbf{.610}&\textbf{.206}&\textbf{.406}&\textbf{.410}&\textbf{.513}&.673&\textbf{.439}&.330&.495&.617&\textbf{.470}&\textbf{.189}&.347&.374&\textbf{.437}\\\hline
		FB15k&HypE-Avg-1t&.567&.229&.141&.312&.387&.223&.081&.155&.157&.258&\textbf{.498}&.216&.135&.269&.319&.185&.179&.261&.152&.258\\
		-237&HypE-Avg-1,2,3t&.567&.295&.197&.351&.444&.224&.109&.185&.218&.292&.492&.277&.185&.300&.366&.185&.233&.309&.206&.291\\
&HypE-Avg&.567&.361&.252&.390&.500&\textbf{.225}&.137&.214&.279&.325&.486&.338&.234&.331&.412&.185&.286&.357&.259&.323\\
&HypE-DS&.549&.361&.247&.398&.519&.208&.135&.234&.277&.324&.473&.341&.233&\textbf{.347}&\textbf{.442}&.174&.285&.393&.257&.327\\
&HypE-TC&\textbf{.573}&\textbf{.366}&\textbf{.256}&\textbf{.399}&\textbf{.528}&\textbf{.225}&\textbf{.146}&\textbf{.246}&\textbf{.283}&\textbf{.335}&.490&\textbf{.344}&\textbf{.238}&.339&.440&\textbf{.186}&\textbf{.306}&\textbf{.411}&\textbf{.260}&\textbf{.335}\\
&HypE (final)&.572&\textbf{.366}&.255&\textbf{.399}&.527&\textbf{.225}&.145&\textbf{.246}&.282&\textbf{.335}&.490&.343&.237&.339&.440&\textbf{.186}&.305&.410&\textbf{.260}&.334\\\hline
		NELL&HypE-Avg-1t&.614&.225&.173&.313&.413&.438&.080&.143&.155&.286&\textbf{.467}&.192&.159&.266&.353&.316&.079&.131&.154&.240\\
		995&HypE-Avg-1,2,3t&.613&.290&.241&.352&.474&.439&.108&.170&.215&.323&.462&.247&.217&.297&.404&.316&.103&.156&.209&.270\\
&HypE-Avg&.612&.354&.309&.391&.534&.440&.135&.197&.275&.359&.456&.302&.275&.328&.455&.316&.127&.180&.263&.299\\
&HypE-DS&.594&.354&.303&.399&.554&.408&.134&.216&.273&.359&.444&.304&.274&\textbf{.344}&\textbf{.488}&.297&.126&.198&.261&.304\\
&HypE-TC&\textbf{.618}&\textbf{.360}&\textbf{.313}&\textbf{.400}&\textbf{.563}&\textbf{.442}&\textbf{.144}&\textbf{.228}&\textbf{.278}&\textbf{.371}&.461&\textbf{.306}&\textbf{.279}&.336&.487&\textbf{.319}&\textbf{.135}&\textbf{.207}&\textbf{.264}&\textbf{.310}\\
&HypE (final)&\textbf{.618}&.359&.312&\textbf{.400}&\textbf{.563}&.441&.143&.227&\textbf{.278}&\textbf{.371}&.460&\textbf{.306}&\textbf{.279}&.336&.486&.318&\textbf{.135}&\textbf{.207}&\textbf{.264}&\textbf{.310}\\\hline
		DBPedia&HypE-Avg-1t&.963&.473&N.A.&.782&.734&.933&.216&.621&.026&.521&.713&.374&N.A.&.589&.540&.696&.170&.467&.020&.390\\
		$|R|=p$&HypE-Avg-1,2,3t&.962&.609&N,A.&.880&.841&.936&.291&.739&.036&.588&.724&.455&N.A.&.658&.619&.696&.221&.553&.027&.438\\
&HypE-Avg&.961&.745&N.A.&.978&.948&.938&.366&.856&.045&.655&.715&.555&N.A.&.726&.697&.696&.271&.638&.034&.486\\
&HypE-DS&.932&.745&N.A.&.997&.985&.869&.362&.938&.045&.654&.697&.560&N.A.&.741&.738&.654&.270&.703&.034&.490\\
&HypE-TC&\textbf{.971}&\textbf{.757}&N.A.&\textbf{1.00}&\textbf{1.00}&\textbf{.941}&\textbf{.388}&\textbf{.986}&\textbf{.047}&\textbf{.677}&\textbf{.722}&\textbf{.563}&N.A.&\textbf{.745}&\textbf{.745}&\textbf{.701}&\textbf{.289}&\textbf{.733}&\textbf{.034}&\textbf{.504}\\
&HypE (final)&.970&.756&N.A.&\textbf{1.00}&\textbf{1.00}&.940&\textbf{.388}&.985&.046&.676&\textbf{.722}&\textbf{.563}&N.A.&.744&.744&.700&\textbf{.289}&\textbf{.733}&\textbf{.034}&.503\\\hline
	\end{tabular}
	\label{tab:ablation_full}
\end{table*}
\begin{table*}[htbp]
	\centering
	\footnotesize	
	\caption{Performance comparison of HypE (ours) against its variants on E-commerce dataset to study the {efficacy} of the model's Query-Search {space}. The columns present the different query structures and averages over them. Best results for each dataset are given in bold.\protect\footnotemark}
	\begin{tabular}{|p{4em}|p{8.1em}|ccc|ccc|ccc|c|ccc|ccc|ccc|c|}
		\hline
		\textbf{Metrics}&&\multicolumn{10}{c|}{\textbf{Hits@3}}&\multicolumn{10}{c|}{\textbf{Mean Reciprocal Rank}}\\\hline
		\textbf{Dataset}&\textbf{Model}&\textbf{1t}&\textbf{2t}&\textbf{3t}& \boldmath$2\cap$ & \boldmath$3\cap$ & \boldmath$2\cup$ & \boldmath$\cap t$ & \boldmath$t\cap$ & \boldmath$\cup t$ & \textbf{Avg}&\textbf{1t}&\textbf{2t}&\textbf{3t}& \boldmath$2\cap$ & \boldmath$3\cap$ &\boldmath$2\cup$&\boldmath$\cap t$&\boldmath$t\cap$&\boldmath$\cup t$&\textbf{Avg}\\\hline
			E-&HypE-Avg-1t&0.0&0.0&0.0&0.0&\textbf{0.0}&0.0&\textbf{0.0}&0.0&0.0&0.0&\textbf{0.0}&0.0&0.0&0.0&\textbf{0.0}&0.0&\textbf{0.0}&0.0&0.0&0.0\\
			commerce&HypE-Avg-1,2,3t&0.0&28.6&39.4&12.5&\textbf{0.0}&0.5&\textbf{0.0}&\textbf{50.0}&39.8&13.0&-1.0&28.5&36.8&13.3&\textbf{0.0}&0.0&\textbf{0.0}&100&35.3&12.6\\
			$|R|=p$&HypE-Avg&-0.3&57.1&77.7&25.0&\textbf{0.0}&0.9&\textbf{0.0}&\textbf{50.0}&78.7&26.0&-2.3&57.0&72.8&23.3&\textbf{0.0}&0.0&\textbf{0.0}&100&\textbf{70.6}&24.2\\
			&HypE-DS&-3.3&57.1&74.5&\textbf{27.5}&\textbf{0.0}&-6.8&\textbf{0.0}&\textbf{50.0}&76.9&26.0&-4.9&58.3&71.9&\textbf{30.0}&\textbf{0.0}&-6.1&\textbf{0.0}&100&69.4&26.3\\
			&HypE-TC&\textbf{0.7}&\textbf{60.0}&\textbf{80.9}&\textbf{27.5}&\textbf{0.0}&\textbf{0.9}&\textbf{0.0}&\textbf{50.0}&\textbf{81.5}&\textbf{31.0}&-1.3&\textbf{59.6}&\textbf{75.4}&\textbf{30.0}&\textbf{0.0}&\textbf{0.6}&\textbf{0.0}&\textbf{200}&\textbf{70.6}&\textbf{28.4}\\
			&HypE (final)&\textbf{0.7}&59.3&79.8&\textbf{27.5}&\textbf{0.0}&\textbf{0.9}&\textbf{0.0}&\textbf{50.0}&80.6&30.0&-1.3&58.9&\textbf{75.4}&26.7&\textbf{0.0}&\textbf{0.6}&\textbf{0.0}&100&\textbf{70.6}&\textbf{28.4}\\\hline
	\end{tabular}
\label{tab:ablation_full_ecomm}
\end{table*}
\addtolength{\tabcolsep}{-1.5pt}    
\begin{table}[htbp]
	\centering
	\footnotesize	
	\caption{Performance comparison of HypE (ours) against the baselines on an e-commerce dataset for intersection and union queries. We fix GQE (E-vector) to be our baseline and report the relative improvements against that for all the methods. The numbers are in percentages. Best results for each dataset are shown in bold.}
	\begin{tabular}{|l|l|ccc|ccc|ccc|ccc|}
		\hline
		\textbf{Metrics}&&\multicolumn{6}{c|}{\textbf{Hits@3}}&\multicolumn{6}{c|}{\textbf{Mean Reciprocal Rank}}\\\hline
		\textbf{Dataset}&\textbf{Model}&\boldmath$2\cap$&\boldmath$3\cap$&\boldmath$2\cup$&\boldmath$\cap t$&\boldmath$t\cap$&\boldmath$\cup t$&\boldmath$2\cap$&\boldmath$3\cap$&\boldmath$2\cup$&\boldmath$\cap t$&\boldmath$t\cap$&\boldmath$\cup t$\\\hline
		$|R|=1$&GQE&0.0&0.0&0.0&0.0&0.0&0.0&\textbf{0.0}&0.0&\textbf{0.0}&\textbf{0.0}&0.0&0.0\\
		&KGAT&\textbf{53.8}&\textbf{2850}&30.8&200&\textbf{300}&0.0&-36.2&-57.3&24.4&-60&-45.5&0.0\\
		&HQE&0.0&0.0&-57.7&0.0&-50.0&\textbf{1500}&-58.5&-99&-61&-80&-90.9&\textbf{1100}\\
		&Q2B&-1.9&-50&32.7&0.0&50&0.0&-59.6&-99&53.7&-80&-81.8&0.0\\
		&HypE&-1.9&-50&\textbf{119.2}&0.0&\textbf{50}&\textbf{1500}&-59.6&-99&\textbf{107.3}&-80&-81.8&\textbf{1100}\\\hline
		$|R|=p$&GQE&\textbf{0.0}&\textbf{0.0}&0.0&\textbf{0.0}&0.0&0.0&0.0&\textbf{0.0}&0.0&0.0&0.0&0.0\\
		&KGAT&\textbf{0.0}&\textbf{0.0}&30.2&\textbf{0.0}&\textbf{2600}&0.0&-8.8&\textbf{0.0}&31.3&-25&0.0&0.0\\
		&HQE&\textbf{0.0}&\textbf{0.0}&0.0&-50&0.0&\textbf{19400}&\textbf{11.8}&\textbf{0.0}&0.0&-75&0.0&\textbf{14400}\\
		&Q2B&\textbf{0.0}&\textbf{0.0}&95.3&\textbf{0.0}&200&0.0&\textbf{11.8}&\textbf{0.0}&59.4&-75&\textbf{100}&0.0\\
		&HypE&\textbf{0.0}&\textbf{0.0}&\textbf{416.3}&\textbf{0.0}&\textbf{2600}&\textbf{19400}&\textbf{11.8}&\textbf{0.0}&\textbf{415.6}&-75&\textbf{100}&\textbf{14400}\\\hline
	\end{tabular}
	\label{tab:ui_query_ecomm}
	\vspace{-1.5em}
\end{table}
\addtolength{\tabcolsep}{-1.5pt}    

\section{Ablation Study: Finer results}
\label{app:ablation}
The finer results, across all datasets, of our Ablation study, described in Section \ref{sec:ablation}, are given in Tables \ref{tab:ablation_full} and \ref{tab:ablation_full_ecomm}. 
We observe that, on average over all types of reasoning queries, a combination of attention aggregation for centers and Deepsets aggregation for limits results in the best performance across all the datasets. Thus, we utilize this combination in our final model.
\section{E-commerce results}
\label{app:e-commerce}
The intersection and union results on the e-commerce datasets, described in Section \ref{sec:query}, are given in Table \ref{tab:ui_query_ecomm}.
\footnotetext{Due to the sensitive nature of the proprietary dataset, we only report relative results.}
\end{document}